\documentclass[10pt,twocolumn,letterpaper]{article}

 \usepackage[pagenumbers]{cvpr} 

\usepackage[dvipsnames]{xcolor}

\usepackage{times}
\usepackage{epsfig}
\usepackage{graphicx}
\usepackage{bm}
\usepackage{mathcommand}
\usepackage{mathrsfs}
\usepackage{amsmath,amsthm,amssymb,mathtools}
\usepackage{times}
\usepackage{epsfig}
\usepackage{graphicx}
\usepackage{amsmath}
\usepackage{amssymb}
\usepackage{bm}
\usepackage{amsthm}
\usepackage{algorithm}
\usepackage{algorithmicx}
\usepackage{algpseudocode}
\usepackage[dvipsnames]{xcolor}
\usepackage{caption}
\usepackage[export]{adjustbox}
\usepackage{float}
\usepackage{stfloats}
\usepackage{mathtools}
\usepackage{tabularx}
\usepackage{booktabs}
\usepackage{float}
\usepackage{placeins}
\usepackage{thm-restate}
\usepackage{thmtools}
\usepackage{wrapfig}
\usepackage{multirow}
\usepackage{enumitem}
\usepackage{colortbl}  
\usepackage{xcolor}
\usepackage{array}   
\usepackage{hhline} 
\usepackage{makecell}
\usepackage{wrapfig}
\usepackage{arydshln}
\newcommand{\bfq}{\mathbf{q}}
\newcommand{\bfk}{\mathbf{k}}
\newcommand{\bfv}{\mathbf{v}}
\newcommand{\bfx}{\mathbf{x}}
\newcommand{\bfo}{\mathbf{o}}
\newcommand{\bfz}{\mathbf{z}}
\newcommand{\bfA}{\mathbf{A}}
\newcommand{\bfB}{\mathbf{B}}
\newcommand{\bfD}{\mathbf{D}}

\newcommand{\bfW}{\mathbf{W}}

\newcommand{\bfM}{\mathbf{M}}

\newcommand{\bff}{\mathbf{f}}

\newcommand{\bfe}{\mathbf{e}}

\newcommand*{\tran}{^{\mkern-1.5mu\mathsf{T}}}

\definecolor{mypink1}{RGB}{241,90,82}
\definecolor{mypink}{RGB}{255,231,226}
\definecolor{myblue}{RGB}{233,250,249}

\definecolor{cvprblue}{rgb}{0.21,0.49,0.74}
\usepackage[pagebackref,breaklinks,colorlinks,citecolor=cvprblue,letterpaper=true,bookmarks=true]{hyperref}

\def\paperID{807} 
\def\confName{CVPR}
\def\confYear{2024}

\title{MACE: Mass Concept Erasure in Diffusion Models}

\author{
	{Shilin Lu}\textsuperscript{1} 
	\quad 
	{Zilan Wang}\textsuperscript{1}
	\quad 
	{Leyang Li}\textsuperscript{1} 
	\quad 
	{Yanzhu Liu}\textsuperscript{2} 
	\quad 
	{Adams Wai-Kin Kong}\textsuperscript{1}\\
	\textsuperscript{1}School of Computer Science and Engineering, Nanyang Technological University, Singapore \\ \hspace{-1cm}\textsuperscript{2}Institute for Infocomm Research ($\rm{I^2R}$) \& Centre for Frontier AI Research (CFAR), A*STAR, Singapore \\
	\hspace{-0.5cm}{\tt\small \{shilin002, wang1982, lile0005\}@e.ntu.edu.sg, liu\_yanzhu@i2r.a-star.edu.sg, adamskong@ntu.edu.sg}
}

\begin{document}
\hypersetup{
	bookmarksnumbered=true,
}

\twocolumn[{
	\renewcommand\twocolumn[1][]{#1}
	\maketitle
	\centering
	\vspace*{-0.9cm}
	\includegraphics[width=1\textwidth]{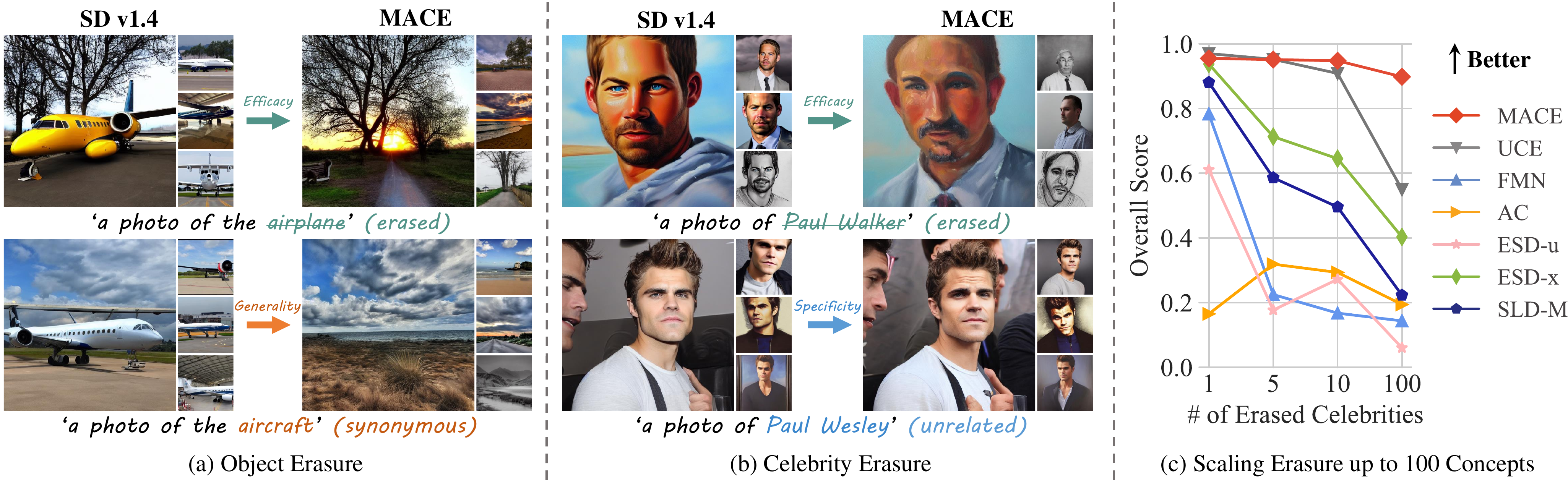} 
	\vspace{-0.7cm}
	\captionof{figure}{Our proposed method, MACE, can erase a large number of concepts from text-to-image diffusion models. This can safeguard celebrity portrait rights, respect copyrights on artworks, and prevent explicit content creation. (a) MACE demonstrates good efficacy and generality by preventing the generation of images reflecting the \textcolor[RGB]{82,148,137}{target concept} and \textcolor[RGB]{197,90,17}{its synonyms}. (b) MACE maintains excellent specificity, ensuring that the \textcolor[RGB]{59,135,205}{unintended concepts} remain intact, even when they share common terms with the \textcolor[RGB]{82,148,137}{target concept}. (c) MACE exhibits a significantly enhanced ability to erase 100 concepts, outperforming previous methods. The overall score indicates the comprehensive erasing capability, as detailed in Section~\ref{sec:exp_cele}.}
	\label{fig:teaser}
	\vspace*{0.2cm}
}]

\begin{abstract}
\vspace{-0.4cm}
The rapid expansion of large-scale text-to-image diffusion models has raised growing concerns regarding their potential misuse in creating harmful or misleading content. In this paper, we introduce MACE, a finetuning framework for the task of MAss Concept Erasure. This task aims to prevent models from generating images that embody unwanted concepts when prompted. Existing concept erasure methods are typically restricted to handling fewer than five concepts simultaneously and struggle to find a balance between erasing concept synonyms (generality) and maintaining unrelated concepts (specificity). In contrast, MACE differs by successfully scaling the erasure scope up to 100 concepts and by achieving an effective balance between generality and specificity. This is achieved by leveraging closed-form cross-attention refinement along with LoRA finetuning, collectively eliminating the information of undesirable concepts. Furthermore, MACE integrates multiple LoRAs without mutual interference. We conduct extensive evaluations of MACE against prior methods across four different tasks: object erasure, celebrity erasure, explicit content erasure, and artistic style erasure. Our results reveal that MACE surpasses prior methods in all evaluated tasks. Code is available at \href{https://github.com/Shilin-LU/MACE}{https://github.com/Shilin-LU/MACE}.
\end{abstract}
\vspace{-0.65cm}
\section{Introduction}
\label{sec:intro}
In large-scale text-to-image (T2I) models \cite{chang2023muse, ding2022cogview2, nichol2021glide, ramesh2022hierarchical, rombach2022high, saharia2022photorealistic, zhou2023pyramid, zhou2024migc, lu2023tf, zhao2023wavelet, zhao2024learning}, the task of concept erasure aims to remove concepts that may be harmful, copyrighted, or offensive. This ensures that when a model is prompted with any phrase related to deleted concepts, it will not generate images reflecting those concepts. 

The drive behind concept erasure is rooted in the significant risks posed by T2I models. These models can generate inappropriate content, such as copyrighted artworks \cite{roose2022art,setty2023suit,art,jiang2023ai}, explicit content \cite{tatum2023porn,schramowski2023safe,zhang2023generate}, and deepfakes \cite{verdoliva2020media,mirsky2021creation}. These issues are largely caused by the unfiltered, web-scraped training data \cite{schuhmann2022laion}. While researchers have put efforts to mitigate these risks through refining datasets and retraining models, these methods are not only costly but also can lead to unforeseen outcomes \cite{carlini2022privacy,connor2022sd}. For example, despite being trained on a sanitized dataset, Stable Diffusion (SD) v2.0 \cite{rombach2022sd2} still produces explicit content. Moreover, it exhibits a diminished generative quality for regular content when compared to its earlier versions  \cite{connor2022sd}. Alternative methods, such as post-generation filtering \cite{dalle32023,rando2022red} and inference guiding \cite{schramowski2023safe,negative_prompt}, are effective when models are accessed only via APIs. Yet, these safeguards can be easily bypassed if users have access to the source code \cite{SmithMano2022}. 

To mitigate the vulnerability of these safeguards, several finetuning-based methods have been proposed \cite{heng2023selective,gandikota2023erasing,gandikota2023unified,kumari2023ablating,zhang2023forget,kim2023towards}. Nonetheless, the challenge of concept erasure lies in balancing the dual requirements of generality and specificity. Generality requires that a concept should be consistently removed, regardless of its expression and the context in which it appears. On the other hand, specificity requires that unrelated concepts remain intact. Our analysis reveals that there is substantial room for enhancing these methods with respect to both generality and specificity. 

\begin{figure}[tbp]
	\centering
	\includegraphics[width=0.98\linewidth]{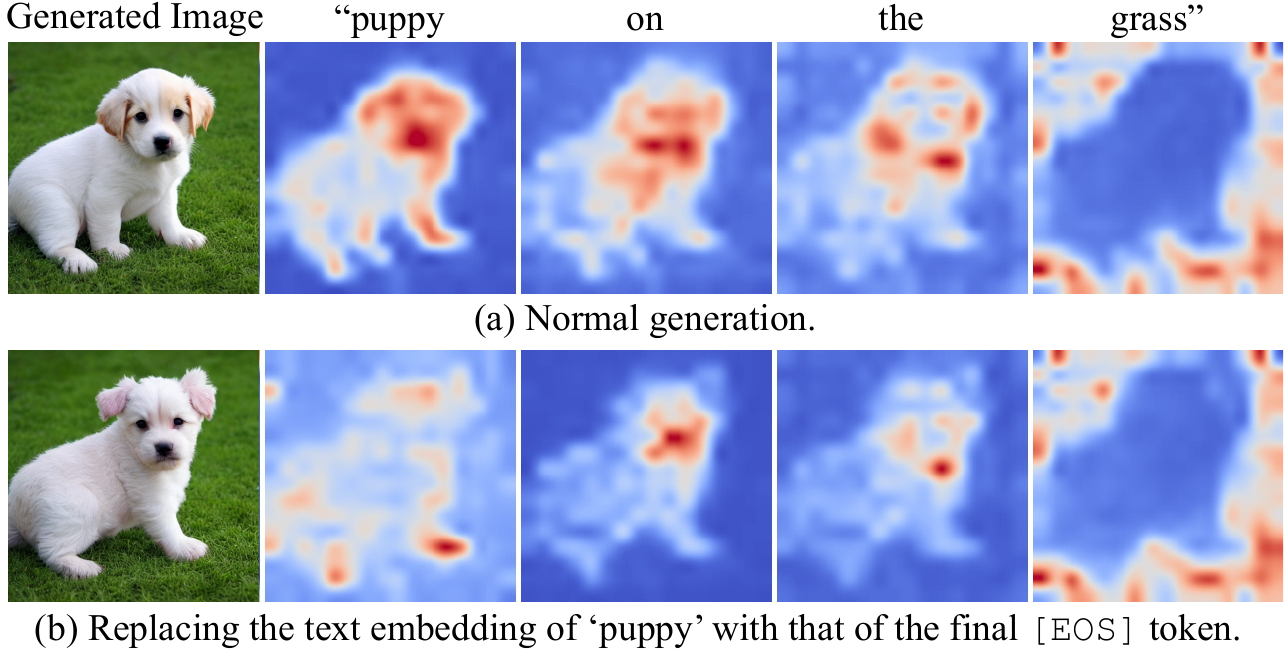}
	\vspace{-0.2cm}
	\caption{\textbf{A concept can be generated solely via residual information:} (a) Average cross-attention map for each word presents that a concept’s information is embedded within other words. (b) A puppy can be generated solely using residual information by replacing the text embedding of `puppy' with that of the final {\tt [EOS]} token. Additional examples are available in Figure~\ref{fig:appendix_residual}.}
	\vspace{-0.2cm}
	\label{fig:evoke}
\end{figure}

We pinpoint three primary issues that hinder the effectiveness of prior works.  Firstly, the information of a phrase is concealed within other words in the prompt through the attention mechanism \cite{vaswani2017attention}. This is sufficient to evoke the concept from T2I models (see Figure~\ref{fig:evoke}), leading to restricted generality and incomplete elimination when removing concepts. 
Secondly, finetuning the diffusion model’s prediction on early denoising steps ($t>t_0$) can result in degraded specificity of concept erasure. Typically, diffusion models generate a general context in the early stage \cite{kwon2022diffusion,raya2023spontaneous,choi2022perception}. For instance, when generating a portrait of Paul Walker or Paul Wesley, the initial sampling trajectory gravitates towards the face manifold. It begins by forming a vague outline that could resemble any person. After a turning point, a.k.a. spontaneous symmetry breaking (SSB) \cite{raya2023spontaneous}, the identity becomes clear with the details progressively filled in. If our goal is to only prevent the model from generating images of Paul Walker, it should not impact other celebrities named `Paul' (See Figure~\ref{fig:teaser}). However, if we alter the predictions made in the early stages, other `Pauls' can inadvertently be affected. Lastly, when fineturning methods are applied to erase a large number of concept (e.g., 100), a noticeable decline in performance is observed. This decline is due to either sequential or parallel finetuning of the models. The former is prone to catastrophic forgetting and the latter results in interference among different concepts being finetuned.

In light of these challenges, we propose a framework, dubbed MAss Concept Erasure (MACE), to erase a large number of concepts from T2I diffusion models. MACE not only achieves a superior balance between generality and specificity, but also adeptly handles the erasure of 100 concepts. It requires neither concept synonyms nor the original training data to perform concept erasure. To remove multiple concepts, MACE starts by refining the cross-attention layers of the pretrained model using a closed-form solution. This design encourages the model to refrain from embedding residual information of the target phrase into other words, thereby erasing traces of the concept in the prompt. Secondly, it employs a unique LoRA module \cite{hu2021lora} for each individual concept to remove its intrinsic information. To maintain specificity, MACE exploits concept-focal importance sampling during LoRA training, mitigating the impact on unintended concepts. Finally, we develop a loss function for MACE to harmoniously integrate multiple LoRA modules without interfering with one another, while preventing catastrophic forgetting. This integration loss can also be swiftly solved using a closed-form solution. We conduct extensive evaluations on four distinct tasks, including object erasure, celebrity erasure, explicit content erasure, and artistic style erasure. MACE demonstrates superior performance on mass concept erasure and strikes an effective balance between specificity and generality, compared with state-of-the-art (SOTA) methods. This achievement paves the way for safer and more regulated T2I applications.

\section{Preliminaries and Related Work}
\label{sec:related}
We provide an overview of two foundational preliminaries upon which our work is based in Appendix~\ref{sec:preliminary}. The first preliminary is the latent diffusion model, which serves as the cornerstone for the T2I model that we investigate. The second one is the cross-attention mechanism, a common feature in T2I models.

~\\
\noindent \textbf{Concept erasure.} Existing research on preventing unwanted outputs from T2I models can be broadly grouped into four categories: post-image filtering \cite{dalle32023,rando2022red}, inference guidance \cite{schramowski2023safe,negative_prompt}, retraining with the curated dataset \cite{rombach2022sd2,nichol2021glide}, and model finetuning \cite{heng2023selective,gandikota2023erasing,gandikota2023unified,kumari2023ablating,zhang2023forget,kim2023towards,ni2023degeneration}. The first two methods are post-hoc solutions and do not address the inherent propensity of the models to generate inappropriate content \cite{SmithMano2022}. Although retraining with curated datasets may offer a solution, it demands significant computational effort and time (e.g., over 150,000 A100 GPU hours for retraining Stable Diffusion) \cite{sd14modelcard}. Finetuning pretrained T2I models is a more viable approach. However, most methods either overlook the residual information of the target phrase embedded within co-existing words, focusing solely on the target phrase \cite{gandikota2023unified,zhang2023forget,gandikota2023erasing}, or they finetune uniformly across timesteps \cite{kumari2023ablating,gandikota2023erasing,zhang2023forget,kim2023towards}. Modifications to diffusion models conditioned on timesteps before SSB \cite{raya2023spontaneous} can negatively affect the generation of retained concepts. In contrast, the proposed MACE addresses these challenges effectively.

~\\
\noindent \textbf{Image cloaking.} An alternative method for safeguarding images against imitation or memorization \cite{somepalli2023diffusion,carlini2023extracting} by T2I models involves an additional step of applying adversarial perturbations to photographs or artworks before they are posted online. This technique, often referred to as cloaking, enables individuals to effectively conceal their images from models during the training phase but remain accessible and discernible to human viewers \cite{salman2023raising,shan2023glaze,zhao2023unlearnable}. Nevertheless, it is crucial to note that this strategy is applicable only to content not yet posted online. To safeguard the vast amount of content already on the web, concept erasure can serve as a viable strategy for large model providers as they prepare to release more advanced models in subsequent evolutions.


\section{Method}
\label{sec:method}
We aim to develop a framework to erase a large number of concepts from pretrained T2I diffusion models. This framework takes two inputs: a pretrained model and a set of target phrases that expresses the concepts to be removed. It returns a finetuned model that is incapable of generating images depicting the concepts targeted for erasing. An effective erasure framework should fulfill the following criteria:

\begin{itemize}[itemsep=0.1em, topsep=0.1em]
	\item \textbf{Efficacy (block target phrases):} If the finetuned model is conditioned on prompts with those target phrases, its outputs should have limited semantic alignment with the prompts. Yet, the outputs should still appear natural, either aligning with a generic category (e.g., sky), or defaulting to the super-category of the concept, if one exists.
	
	\item \textbf{Generality (block synonyms):} The model should also prevent the generation of images semantically related to any synonyms of the targeted phrases, ensuring that the erasure is not limited to the exact wording of the prompts.
	
	\item \textbf{Specificity (preserve unrelated concepts):} If the finetuned model is conditioned on prompts that are semantically unrelated to the erased concepts, its output distribution should closely align with that of the original model.
\end{itemize}

\begin{figure}[tbp]
	\centering
	\includegraphics[width=1\linewidth]{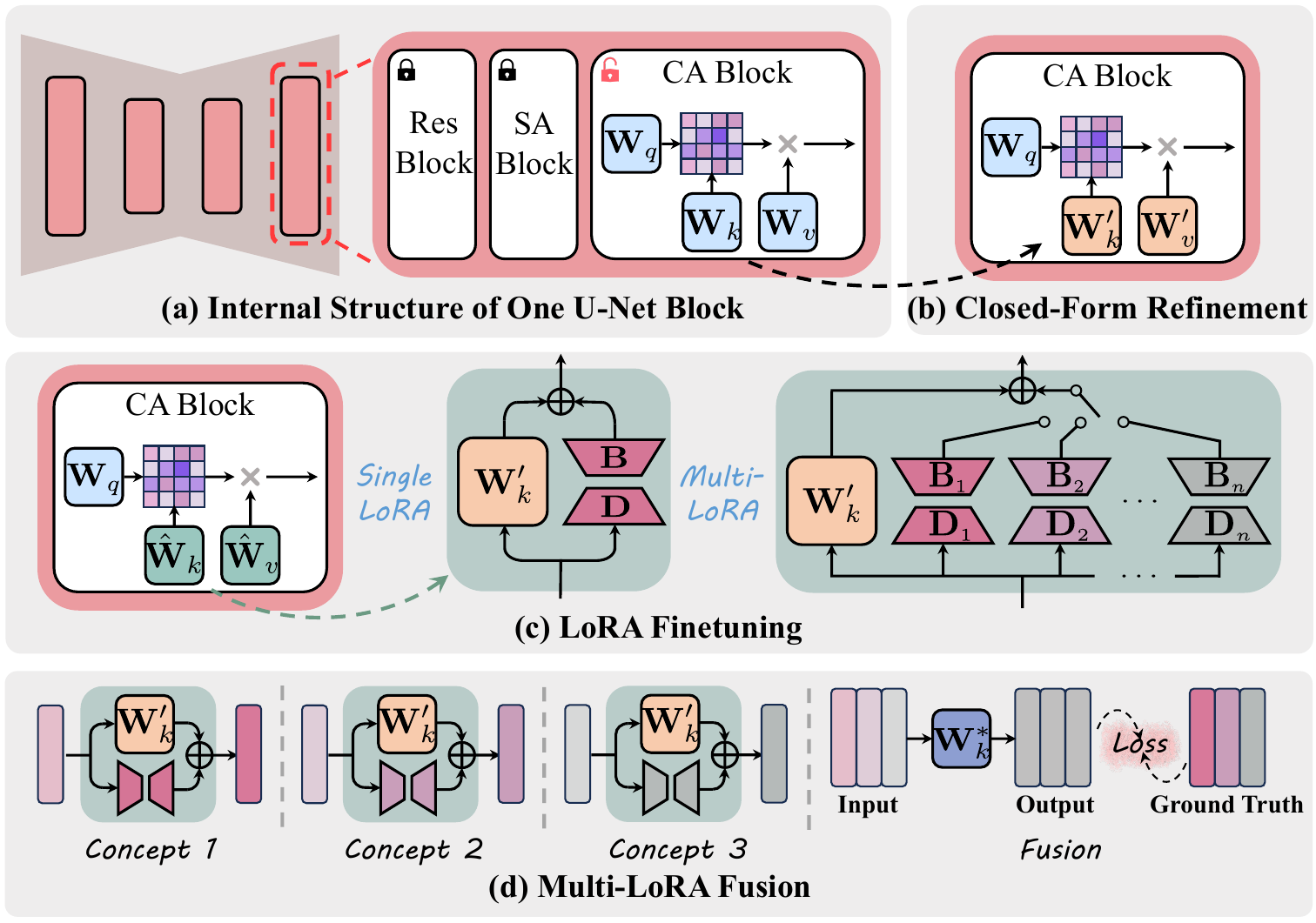}
	\vspace{-0.4cm}
	\caption{\textbf{Overview of MACE:} (a) Our framework focuses on tuning the prompts-related projection matrices, \(\bfW_k\) and \(\bfW_v\), within cross-attention (CA) blocks. (b) (Section~\ref{sec:close} \& Figure~\ref{fig:close}) The pretrained U-Net's CA blocks are refined using a closed-form solution, discouraging the model from embedding the residual information of the target phrase into surrounding words. (c) (Section~\ref{sec:single} \& Figure~\ref{fig:lora}) For each concept targeted for removal, a distinct LoRA module is learned to eliminate its intrinsic information. (d) (Section~\ref{sec:fusion}) A closed-form solution is introduced to integrate multiple LoRA modules without interfering with one another while averting catastrophic forgetting.}
	\vspace{-0.2cm}
	\label{fig:overview}
\end{figure}

To this end, we introduce MACE, a MAss Concept Erasure framework. The information of a phrase is embedded not only within the phrase itself but also within the words it co-exists with. To effectively erase the targeted concepts, our framework first removes the residual information from the co-existing words (Section~\ref{sec:close}). Subsequently, distinct LoRA modules are trained to eliminate the intrinsic information specific to each target concept (Section~\ref{sec:single}). Lastly, our framework integrates multiple LoRA modules without mutual interference, leading to a final model that effectively forgets a wide array of concepts (Section~\ref{sec:fusion}). Figure~\ref{fig:overview} presents an overview of our framework.

\subsection{Closed-Form Cross-Attention Refinement}
\label{sec:close}
In this section, we suggest a closed-form cross-attention refinement to encourage the model to refrain from embedding residual information of the target phrase into other words. Such residual information is adequate to evoke the unwanted concept from T2I models. The root of this issue lies in the attention mechanism \cite{vaswani2017attention}, where the text embedding of a token encapsulates information from other tokens. This results in its `Key' and `Value' vectors absorbing and reflecting information from other tokens. 

\begin{figure}[tbp]
	\centering
	\vspace{-0.2cm}
	\includegraphics[width=1\linewidth]{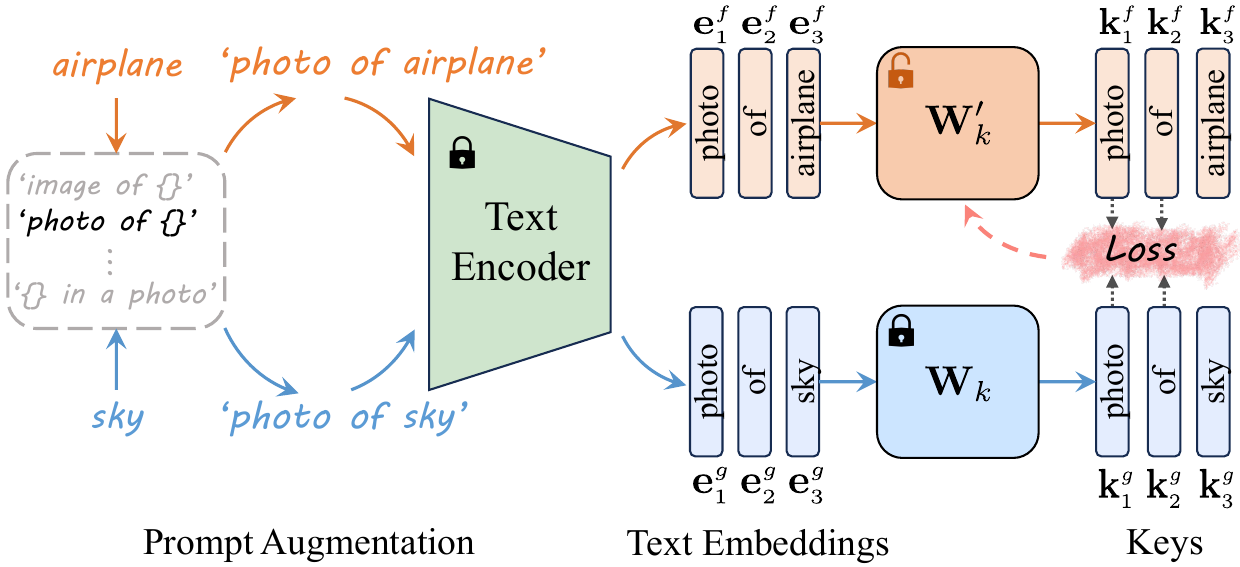}
	\vspace{-0.5cm}
	\caption{\textbf{Closed-Form Cross-Attention Refinement:} The $\bfW_k^\prime$ is tuned such that the `Keys’ of words co-existing with the target phrase `airplane’ are mapped to the `Keys’ of those same words when the target phrase is replaced with a generic concept ‘sky’.}
	\vspace{-0.3cm}
	\label{fig:close}
\end{figure}

To tackle this, we focus on refining the cross-attention modules, which play a pivotal role in processing text prompts. For example, when altering the projection matrix $\bfW_k$, we modify it such that the `Keys' of the words that co-exist with the target phrase in the prompt are mapped to the `Keys' of those same words in another prompt, where the target phrase is replaced with either its super-category or a generic concept. Notably, the `Keys' of the target phrase itself remain unchanged to avoid impacting on other unintended concepts associated with that phrase. Figure~\ref{fig:close} illustrates this process using the projection matrix $\bfW_k$, and the same principle is applicable to $\bfW_v$.

Drawing upon methods that view matrices as linear associative memories \cite{anderson1972simple,kohonen2012associative}, often used to edit knowledge embedded within neural networks \cite{arad2023refact,meng2022locating,meng2022mass,bau2020rewriting,orgad2023editing,gandikota2023unified,basu2023localizing}, we formulate our objective function as follows:
\begin{equation} 
	\label{eq:close-form-loss}
	\begin{split}
			 &\min\limits_{\bfW^{\prime}_k} \sum\limits_{i=1}^n \left\| \bfW_k^{\prime} \cdot \bfe^f_i - \bfW_k \cdot \bfe_i^g  \right\|_2^2 \\
		&+  \lambda_1 \sum\limits_{i=n+1}^{n+m} \left\| \bfW_k^{\prime} \cdot \bfe^p_i - \bfW_k \cdot \bfe_i^p  \right\|_2^2,
	\end{split}
\end{equation}
where $\lambda_1 \in \mathbb{R}^{+}$ is a hyperparameter, $\bfe_i^f$ is the embedding of a word co-existing with the target phrase, $\bfe_i^g$ is the embedding of that word when the target phrase is replaced with its super-category or a generic one, $\bfe_i^p$ is the embedding for preserving the prior, $\bfW_k$ is the pretrained weights, and $n,m$ are the number of embeddings for mapping and preserving, respectively. As derived in Appendix~\ref{sec:proof}, this optimization problem has a closed-form solution:
\begin{align} 
		\bfW_k^\prime = & \left( \sum\limits_{i=1}^n \bfW_k \cdot \bfe^g_i \cdot (\bfe_i^f)\tran + \lambda_1 \sum\limits_{i=n+1}^{n+m} \bfW_k \cdot \bfe^p_i \cdot (\bfe_i^p)\tran \right) \nonumber \\
		\cdot & \left( \sum\limits_{i=1}^n \bfe^f_i \cdot (\bfe^f_i)\tran + \lambda_1 \sum\limits_{i=n+1}^{n+m} \bfe^p_i \cdot (\bfe^p_i)\tran \right)^{-1},
	\label{eq:close-form-solution}
\end{align}
where $\sum_{i=n+1}^{n+m} \bfW_k  \bfe^p_i  (\bfe_i^p)\tran$ and $\sum _{i=n+1}^{n+m} \bfe^p_i (\bfe^p_i)\tran$ are pre-cached constants for preserving prior. These constants are capable of encapsulating both general and domain-specific knowledge, as detailed in Appendix~\ref{sec:proof}. The general knowledge is estimated on the MS-COCO dataset \cite{lin2014microsoft} by default. 

\subsection{Target Concept Erasure with LoRA}
\label{sec:single}

After applying the closed-form refinement to eliminate the traces of the target concepts from co-existing words (Section \ref{sec:close}), our focus shifts to erasing the intrinsic information within the target phrase itself. 

\begin{figure}[tbp]
	\centering
		\vspace{-0.2cm}
	\includegraphics[width=1\linewidth]{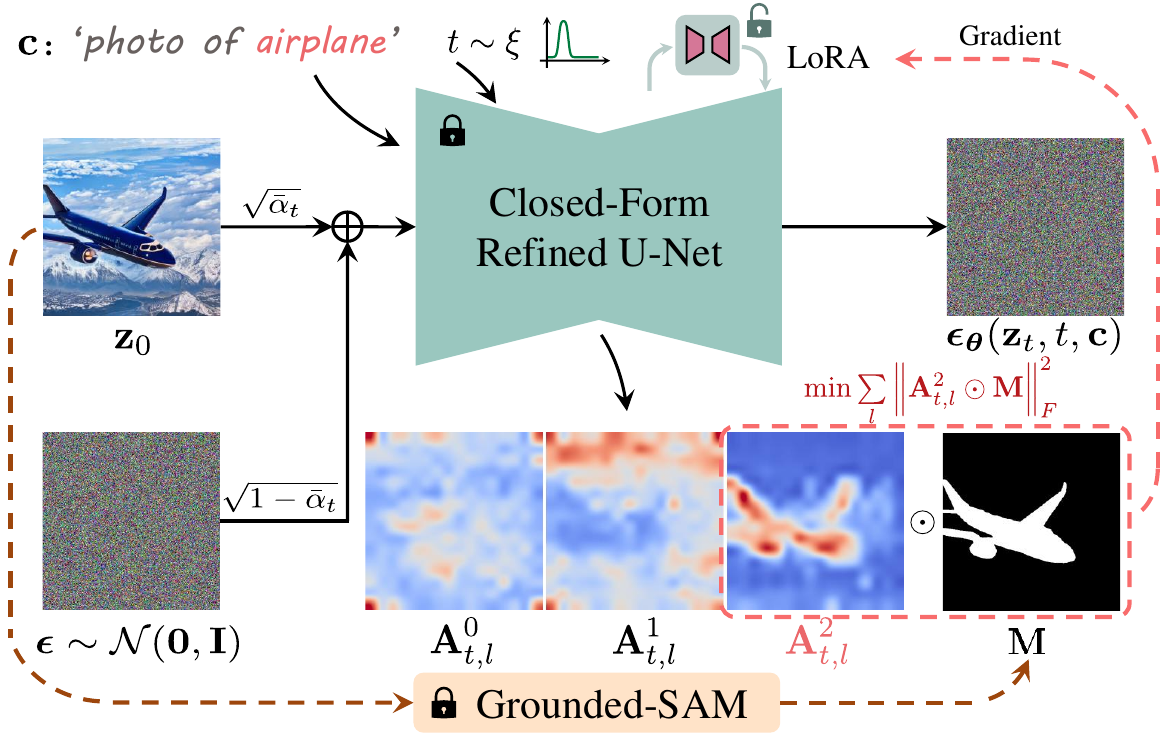}
		\vspace{-0.5cm}
	\caption{\textbf{Training with LoRA to Erase Intrinsic Information:} Eight images are generated for each target concept as a training set via SD v1.4. To obtain the attention maps, the images undergo forward diffusion to timestep $t$ and then are fed into the closed-form refined model for predicting noise at timestep $t$. The LoRA modules are trained to reduce the activation in the masked attention maps that correspond to the target phrase.}
		\vspace{-0.3cm}
	\label{fig:lora}
\end{figure}

~\\
\noindent \textbf{Loss function.} Intuitively, if a concept is to appear in generated images, it should exert significant influence on several patches of those images \cite{chefer2023attend,phung2023grounded}. This implies that the attention maps corresponding to the tokens of the concept should display high activation values in certain regions. We adapt this principle in an inverse manner to eliminate the information within the target phrase itself. The loss function is designed to suppress the activation in certain regions of the attention maps that correspond to the target phrase tokens. These specific regions are identified by segmenting the input image with Grounded-SAM \cite{kirillov2023segany,liu2023grounding}. Figure~\ref{fig:lora} depicts the training process. The loss function is defined as:
\begin{align} 
	\min \sum\limits_{i \in S} \sum\limits_{l}\left\| \bfA_{t,l}^i \odot \bfM \right\|^2_F,
	\label{eq:attn}
\end{align}
where $S$ is the set of indices corresponding to the tokens of the target phrase, $\bfA_{t,l}^i$ is the attention map of token $i$ at layer $l$ and timestep $t$, $\bfM$ is the segmentation mask, and $\lVert\cdot\rVert_F$ is the Frobenius norm.

~\\
\noindent \textbf{Parameter subset to finetune.} 
To minimize the loss function (Eq.~(\ref{eq:attn})), we tune the closed-form refined projection matrices, $\bfW_k^\prime$ and $\bfW_v^\prime$, by identifying a set of weight modulations, $\Delta \bfW_k$ and $\Delta \bfW_v$. Determining high-dimensional modulation matrices in large-scale models is non-trivial. However, weight modulations usually have a low intrinsic rank when they are adapted for specific downstream tasks \cite{hu2021lora}. Hence, we decompose the modulation matrices using LoRA \cite{hu2021lora}. Specifically, for each target concept and each projection matrix (e.g., $\bfW_k^{\prime} \in \mathbb{R}^{d_\text{in} \times d_\text{out}}$), we learn two matrices, $\bfB \in \mathbb{R}^{d_\text{in} \times r}$ and $\bfD \in \mathbb{R}^{r \times d_\text{out}}$, where $r \ll \min(d_\text{in}, d_\text{out})$ is the decomposition rank. The new modulated matrices are:
\begin{align} 
	\hat{\bfW}_k = \bfW_k^{\prime} + \Delta \bfW_k = \bfW_k^{\prime} + \bfB \times \bfD.
	\label{eq:lora}
\end{align}

\noindent \textbf{Concept-focal importance sampling (CFIS).} 
If the attention loss (Eq.~(\ref{eq:attn})) is computed based on attention maps that are obtained at uniformly sampled timesteps, the predicted score function at various noise levels will be affected. Consequently, it will influence the entire sampling trajectory, undermining the specificity of concept erasure. This issue is especially problematic when erasing phrases that contain polysemous words or common surnames and given names. The reason lies in the nature of the diffusion trajectory. The sample initially gravitates towards the data manifold and possesses the potential to converge to various concept modes associated with the conditional phrase \cite{chen2023geometric,raya2023spontaneous}. After the turning point (a.k.a., SSB \cite{raya2023spontaneous}), the specific mode to be fully denoised is determined \cite{raya2023spontaneous}. Our goal is to influence only the path leading to a particular mode, such as ‘Bill Clinton’, rather than affecting paths leading to every celebrity named ‘Clinton’ or ‘Bill’. Thus, it is crucial that the early sampling trajectory remains largely unaffected.

To this end, we opt not to sample the timestep $t$ from a uniform distribution when training LoRA. Instead, we introduce a sampling distribution that assigns greater probability to smaller values of $t$. The probability density function for sampling $t$ is defined as (A graph of this function is provided in Appendix~\ref{sec:CFIS_pdf}):
\begin{align} 
	\xi(t) = \frac{1}{Z} \left( \sigma\left( \gamma(t-t_1) \right) - \sigma\left( \gamma(t-t_2) \right) \right),
	\label{eq:sampling}
\end{align}
where $Z$ is a normalizer, $\sigma(x)$ is the sigmoid function $1/(1+e^{-x})$, with $t_1$ and $t_2$ as the bounds of a high probability sampling interval $(t_1 < t_2)$, and $\gamma$ as a temperature hyperparameter. We empirically set $t_1=200, t_2=400$, and $\gamma = 0.05$ throughout our experiments. 

This strategy excels particularly in eliminating mass concepts that share overlapping terms. However, it is important to note that when erasing a smaller number of concepts or mass concepts that do not share overlapping terms, the improvements tend to be incremental. Beyond enhancing specificity, this design significantly boosts training effectiveness by fostering a more concentrated and efficient learning process.

\subsection{Fusion of Multi-LoRA Modules}
\label{sec:fusion}
In this section, we present a scheme to fuse multiple LoRA modules. Each LoRA module acts as a conceptual suppressor for the pretrained model, inducing a state of amnesia wherein the model loses their grasp on a specific concept. When working collaboratively, these modules should collectively enable the model to forget all the concepts targeted for erasure. A naïve solution for integrating LoRA modules is to utilize a weighted sum \cite{lora_stable}:
\begin{align} 
	\hat{\bfW}_{k} = \bfW_{k}^\prime +\sum\limits_{i=1}^q \omega_i \Delta \bfW_{k,i}, ~~~~\text{s.t.} \sum\limits_{i=1}^q \omega_i = 1,
\end{align}
where $\bfW_k^\prime$ is the closed-form refined weight, $\Delta \bfW_{k,i}$ is the LoRA module associated with the $i$th concept, $\omega_i$ is the normalized weighting factor, and $q$ is the number of the target concepts. This naïve fusion method leads to interference among the modules, thereby diminishing the erasure performance, as evidenced in the ablation study (Section~\ref{sec:exp_abl}). 

To preserve the capability of LoRA modules, we introduce a novel fusion technique illustrated in Figure~\ref{fig:overview}~(d). We input the text embeddings of the target phrases into the respective LoRA module. The resulting outputs serve as the ground truth for optimizing the projection matrices. The objective function is defined by:
\begin{equation} 
	\label{eq:close-form-fuse}
	\begin{split}
		 \min\limits_{\bfW_k^*} &\sum\limits_{i=1}^q \sum\limits_{j=1}^p \left\|  \bfW_k^* \cdot \bfe_j^f  - \left(\bfW_{k}^\prime+ \Delta \bfW_{k,i}\right) \cdot \bfe^f_j \right\|_2^2 \\
		& +  \lambda_2 \sum\limits_{j=p+1}^{p+m} \left\| \bfW_k^* \cdot \bfe_j^p - \bfW_k \cdot \bfe^p_j  \right\|_2^2,
	\end{split}
\end{equation}
where $\bfW_{k}$ is the original weight, $\bfW_{k}^\prime$ is the closed-form refined weight, $\bfe_i^f$ is the embedding of a word co-existing with the target phrase, $\bfe_j^p$ is the embedding for prior preserving, $\lambda_2 \in \mathbb{R}^{+}$ is a hyperparameter, $q$ is the number of erased concepts, and $p,m$ are the number of embeddings for mapping and preserving. Similar to Eq.~(\ref{eq:close-form-solution}), this optimization problem has a closed-form solution as well.

Compared with sequential or parallel finetuning of a pretrained model for erasing multiple concepts, employing separate LoRA modules for each concept and then integrating them offers better prevention against catastrophic forgetting and provides more flexibility.

\section{Experiments}
In this section, we conduct a comprehensive evaluation of our proposed method, benchmarking it against SOTA baselines across four tasks. The baselines comprise ESD-u \cite{gandikota2023erasing}, ESD-x \cite{gandikota2023erasing}, FMN \cite{zhang2023forget}, SLD-M \cite{schramowski2023safe}, UCE \cite{gandikota2023unified}, and AC \cite{kumari2023ablating}. The four tasks are: object erasure (Section~\ref{sec:exp_object}), celebrity erasure (Section~\ref{sec:exp_cele}), explicit content erasure (Section~\ref{sec:exp_explicit}), and artistic style erasure (Section~\ref{sec:exp_art}). All results from the original SD~v1.4 and SD~v2.1 are obtained without the application of negative prompts.

Our evaluation not only measures efficacy but also explores the generality and specificity of the erasure methods. The generality assessment is primarily conducted in the object erasure, since synonyms for a particular object tend to be precise and universally acknowledged compared to those for celebrities and artists. Evaluating specificity is more straightforward and is therefore applied across all tasks. We also focus on the effectiveness of these methods in handling multi-concept erasure, using the celebrity erasure as a key benchmark. We then highlight the superior performance of our proposed method in erasing explicit content and artistic styles. Lastly, we conduct ablation studies (Section~\ref{sec:exp_abl}) to understand the impact of the key components.

\subsection{Implementation Details}
\label{sec:lmple}
We finetune all models on SD v1.4 and generate images with DDIM sampler \cite{song2020denoising} over 50 steps. We follow  \cite{kumari2023ablating} to augment the input target concept using prompts generated by the GPT-4 \cite{gpt2023}. The prompt augmentation varies depending on the target concept type (e.g., objects or styles). Each LoRA module is trained for 50 gradient update steps. We implement baselines as per the configurations recommended in their original settings. Further details are provided in Appendix~\ref{sec:imple_details}.

\begin{table*}[tbp]
	\caption{\textbf{Evaluation of Erasing the CIFAR-10 Classes:} Results for the first four individual classes, along with the average results across 10 classes, are presented. CLIP classification accuracies are reported for each erased class in three sets: the erased class itself ($\text{Acc}_\text{e}$, efficacy), the nine remaining unaffected classes ($\text{Acc}_\text{s}$, specificity), and three synonyms of the erased class ($\text{Acc}_\text{g}$, generality). The \textcolor{mypink1}{harmonic means} $H_\text{o}$ reflect the comprehensive erasure capability. All presented values are denoted in percentage (\%). Results pertaining to the latter six classes are available in Appendix~\ref{sec:additional_cifar_10}. The classification accuracies of images generated by the original SD v1.4 are presented for reference.}
	\vspace{-0.4cm}
	\begin{center}
		\resizebox{1\textwidth}{!}{
			\begin{tabular}{lccc>{\columncolor{mypink}}cccc>{\columncolor{mypink}}cccc>{\columncolor{mypink}}cccc>{\columncolor{mypink}}cccc>{\columncolor{mypink}}c}
				\toprule
				\multirow{2}{*}{Method} & \multicolumn{4}{c}{Airplane Erased} & \multicolumn{4}{c}{Automobile Erased} & \multicolumn{4}{c}{Bird Erased} & \multicolumn{4}{c}{Cat Erased} & \multicolumn{4}{c}{\textbf{Average across 10 Classes}} \\
				\cmidrule(lr){2-5}  \cmidrule(lr){6-9} \cmidrule(lr){10-13} \cmidrule(lr){14-17} \cmidrule(lr){18-21}  
				& $\text{Acc}_\text{e}$ $\downarrow$ & $\text{Acc}_\text{s}$ $\uparrow$ & $\text{Acc}_\text{g}$ $\downarrow$ & ${H_\text{o} }$ $\uparrow$ & $\text{Acc}_\text{e}$ $\downarrow$ & $\text{Acc}_\text{s}$ $\uparrow$ & $\text{Acc}_\text{g}$ $\downarrow$ & ${H_\text{o} }$ $\uparrow$ & $\text{Acc}_\text{e}$ $\downarrow$ & $\text{Acc}_\text{s}$ $\uparrow$ & $\text{Acc}_\text{g}$ $\downarrow$ & ${H_\text{o} }$ $\uparrow$ & $\text{Acc}_\text{e}$ $\downarrow$ & $\text{Acc}_\text{s}$ $\uparrow$ & $\text{Acc}_\text{g}$ $\downarrow$ & ${H_\text{o} }$ $\uparrow$ & $\text{Acc}_\text{e}$ $\downarrow$ & $\text{Acc}_\text{s}$ $\uparrow$ & $\text{Acc}_\text{g}$ $\downarrow$ & ${H_\text{o} }$ $\uparrow$  \\
				\midrule
				FMN \cite{zhang2023forget} & 96.76 & 98.32 & 94.15 & 6.13 & 95.08  & 96.86 & 79.45 & 11.44 & 99.46 & 98.13 & 96.75 & 1.38 & 94.89 & 97.97 & 95.71 & 6.83 & 96.96 & 96.73 & 82.56 & 6.13 \\
				AC \cite{kumari2023ablating} & 96.24 & 98.55 & 93.35 & 6.11 & 94.41 & 98.47 & 73.92 & 13.19 & 99.55 & 98.53 & 94.57 & 1.24 & 98.94 & 98.63 & 99.10 & 1.45 & 98.34 & 98.56 & 83.38 & 3.63 \\
				UCE \cite{gandikota2023unified} & 40.32 & 98.79 & 49.83 & 64.09 & 4.73 & 99.02 & 37.25 & 82.12 & 10.71 & 98.35 & 15.97 & 90.18 & 2.35 & 98.02 & 2.58 & \textbf{97.70} & 13.54 & 98.45 & 23.18 & 85.48 \\
				SLD-M \cite{schramowski2023safe} & 91.37 & 98.86 & 89.26 & 13.69 & 84.89 & 98.86 & 66.15 & 28.34 & 80.72 & 98.39 & 85.00 & 23.31 & 88.56 & 98.43 & 92.17 & 13.31 & 84.14 & 98.54 & 67.35 & 26.32 \\
				ESD-x \cite{gandikota2023erasing} & 33.11 & 97.15 & 32.28 & 74.98 & 59.68 & 98.39 & 58.83 & 50.62 & 18.57 & 97.24 & 40.55 & 76.17 & 12.51 & 97.52 & 21.91 & 86.98 & 26.93 & 97.32 & 31.61 & 76.91 \\
				ESD-u \cite{gandikota2023erasing} & 7.38 & 85.48 & 5.92 & 90.57 & 30.29 & 91.02 & 32.12 & 74.88 & 13.17 & 86.17 & 20.65 & 83.98 & 11.77 & 91.45 & 13.50 & 88.68 & 18.27 & 86.76 & 16.26 & 83.69 \\
				Ours & 9.06 & 95.39 & 10.03 & \textbf{92.03} & 6.97 & 95.18 & 14.22 & \textbf{91.15} & 9.88 & 97.45 & 15.48 & \textbf{90.39} & 2.22 & 98.85 & 3.91 & 97.56 & 8.49 & 97.35 & 10.53 & \textbf{92.61} \\
				\midrule
				\rowcolor{white} 
				SD v1.4 \cite{rombach2022high} & 96.06 & 98.92 & 95.08 & - & 95.75  & 98.95 & 75.91 & - & 99.72 & 98.51 & 95.45 & - & 98.93 &98.60 & 99.05 & - & 98.63 & 98.63 & 83.64 & - \\
				\bottomrule
		\end{tabular}}
	\end{center}
		\vspace{-0.5cm} 
	\label{tab:object}
\end{table*}

\subsection{Object Erasure}
\label{sec:exp_object}

\noindent \textbf{Evaluation setup.} For each erasure method, we finetune ten models, with each model designed to erase one object class of the CIFAR-10 dataset \cite{krizhevsky2009learning}. To assess erasure efficacy, we use each finetuned model to generate 200 images of the intended erased object class, prompted by \textit{`a photo of the \{erased class name\}'}. These images are classified using CLIP \cite{radford2021learning}, and the criterion for successful erasure is a low classification accuracy. To assess specificity, we use each finetuned model to generate 200 images for each of the nine remaining, unmodified object classes with prompts \textit{`a photo of the \{unaltered class name\}'}. A high classification accuracy indicates excellent erasure specificity. For assessing generality, we prepare three synonyms for each object class, listed in Table~\ref{tab:appendix_syn}. Each finetuned model is used to generate 200 images for each synonym associated with the erased class, using the prompt \textit{`a photo of the \{synonym of erased class name\}'}. In this case, good generality is reflected by lower classification accuracies.

Importantly, to evaluate the overall erasure capability of methods, we use the harmonic mean of efficacy, specificity, and generality. It is calculated as follows:
\begin{align} 
	H_\text{o} = \frac{3}{ \left(1 - \text{Acc}_\text{e}\right)^{-1} +\left(\text{Acc}_\text{s}\right)^{-1} + \left(1 - \text{Acc}_\text{g}\right)^{-1} },
	\label{eq:harmonic}
\end{align}
where $H_\text{o}$ is the harmonic mean for object erasure, $\text{Acc}_\text{e}$ is the accuracy for the erased object (efficacy), $\text{Acc}_\text{s}$ for the remaining objects (specificity), and $\text{Acc}_\text{g}$ for the synonyms of the erased object (generality). A lower value of $\text{Acc}_\text{e}$ and $\text{Acc}_\text{g}$, and a higher $\text{Acc}_\text{s}$ contribute to a higher harmonic mean, indicating a superior comprehensive erasure ability. 

~\\
\noindent \textbf{Discussions and analysis.} Table~\ref{tab:object} presents the results of erasing the first four object classes of the CIFAR-10 dataset,  as well as the average results across all 10 classes. The results of the latter six classes can be found in Appendix~\ref{sec:additional_cifar_10}. Our approach attains the highest harmonic mean across the erasure of nine object classes, with the exception of `cat', where our performance nearly matches the top result. This underscores the superior erasure capabilities of our approach, striking an effective balance between specificity and generality. Additionally, it is noteworthy that while methods like FMN \cite{zhang2023forget} and AC \cite{kumari2023ablating} are proficient in removing specific features of a subject, they fall short in completely eradicating the subject’s generation.

\subsection{Celebrity Erasure}
\label{sec:exp_cele}

\noindent \textbf{Evaluation setup.} In this section, we evaluate the erasure methods with respect to their ability to erase multiple concepts. We establish a dataset consisting of 200 celebrities whose portraits, generated by SD v1.4, are recognizable with remarkable accuracy $(>99\%)$ by the GIPHY Celebrity Detector (GCD) \cite{GCD}. The dataset is divided into two groups: an erasure group with 100 celebrities whom users aim to erase, and a retention group with 100 other celebrities whom users intend to preserve. The complete list of these celebrities is provided in Table~\ref{tab:appendix_cele}.

We perform a series of four experiments where SD v1.4 is finetuned to erase 1, 5, 10, and all 100 celebrities in the erasure group. The efficacy of each erasure method is tested by generating images of the celebrities intended for erasure. Successful erasure is measured by a low top-1 GCD accuracy in correctly identifying the erased celebrities. To test the specificity of methods on the retained celebrities, we generate and evaluate images of the celebrities in the retention group in the same way. A high specificity is indicated by a high top-1 GCD accuracy.

Similar to Eq.~(\ref{eq:harmonic}), we underscore the comprehensive ability of the multi-concept erasure method by computing the harmonic mean of efficacy and specificity:
\begin{align} 
	H_\text{c} = \frac{2}{ \left(1 - \text{Acc}_\text{e}\right)^{-1} +\left(\text{Acc}_\text{s}\right)^{-1} },
	\label{eq:harmonic1}
\end{align}
where $H_c$ is the harmonic mean for celebrity erasure, $\text{Acc}_\text{e}$ is the accuracy for the erased celebrities (efficacy), and $\text{Acc}_\text{s}$ for the retained celebrities (specificity). Furthermore, we assess the specificity of methods on regular content utilizing the MS-COCO dataset~\cite{lin2014microsoft}. We sample 30,000 captions from the validation set to generate images and evaluate FID~\cite{parmar2022aliased} and CLIP score~\cite{radford2021learning}.

\begin{figure*}[tbp]
	\centering
	\includegraphics[width=1\linewidth]{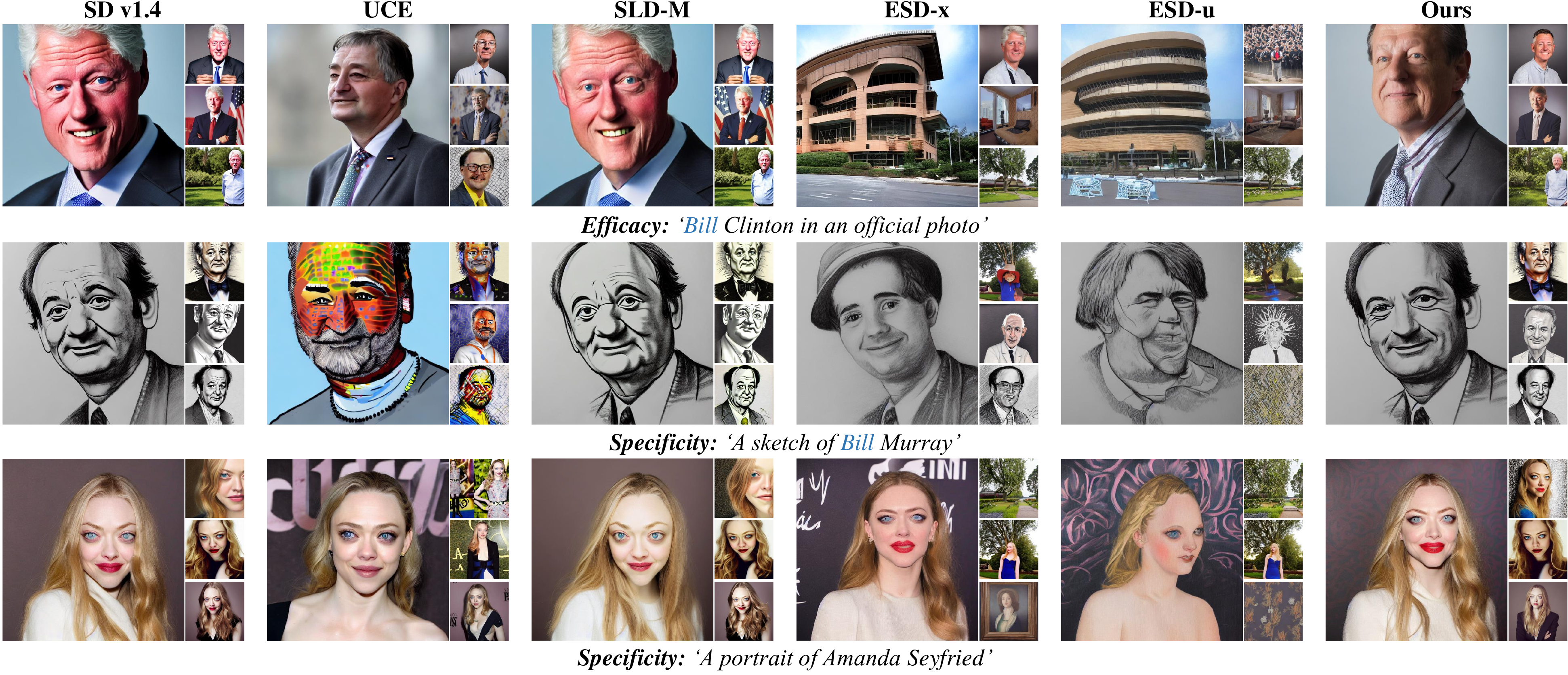}
	\vspace{-0.5cm}
	\caption{\textbf{Qualitative Comparison of Erasing 100 Celebrities from SD v1.4:} Bill Clinton belongs to the erasure group for assessing efficacy, while Bill Murray and Amanda Seyfried are in the retention group to evaluate specificity. Preserving Bill Murray's images is challenging, as his first name is the same as Bill Clinton's, who is in the erasure group. Additional examples are in Appendix~\ref{sec:addtional_qualitative}.}
	\vspace{-0.3cm}
	\label{fig:cele_erasure}
\end{figure*}

\begin{figure}[tbp]
	\centering
	\includegraphics[width=1\linewidth]{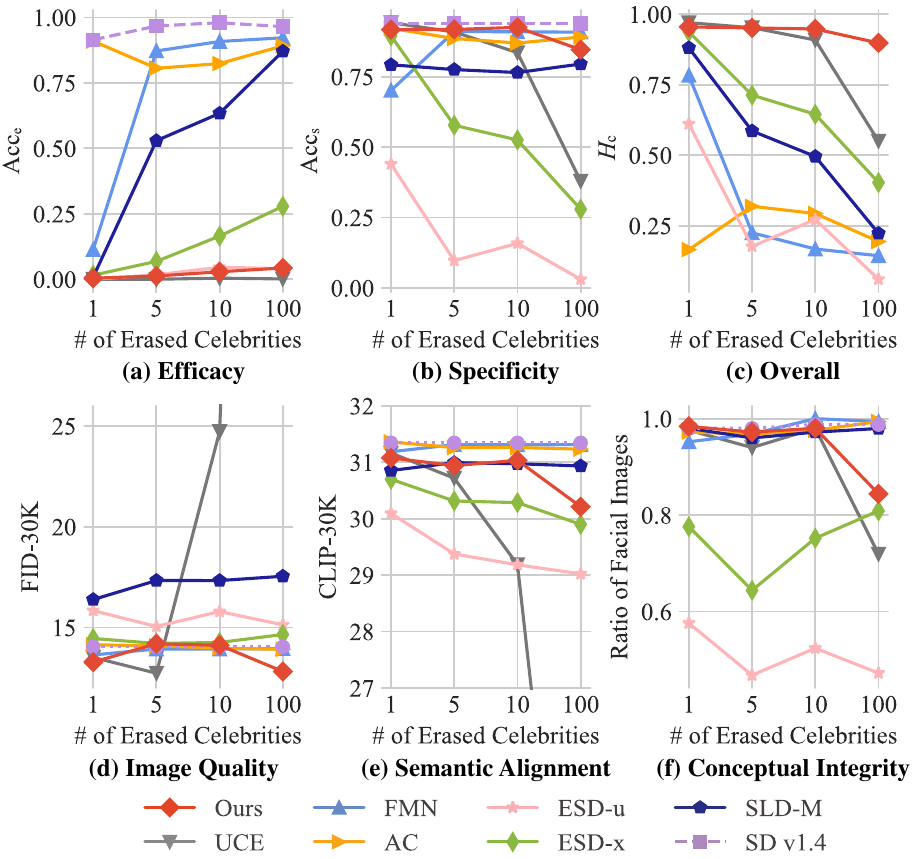}
	\vspace{-0.5cm}
	\caption{\textbf{Evaluation of Erasing Multiple Celebrity:} The evaluation metrics include the detection accuracy on images of erased celebrities ($\text{Acc}_\text{e} \downarrow$) and those of retained celebrities ($\text{Acc}_\text{s} \uparrow$), the harmonic mean ($H_\text{c} \uparrow $) which indicates overall erasure performance, FID, CLIP score, and the ratio of facial images.}
		\vspace{-0.2cm}
	\label{fig:cele_curve}
\end{figure}

~\\
\noindent \textbf{Discussions and analysis.} Figure~\ref{fig:cele_curve}~(c) illustrates a notable enhancement in overall erasure performance achieved by our method, particularly when 100 concepts are erased. This improvement indicates a more effective balance between efficacy and specificity. FMN~\cite{zhang2023forget}, AC~\cite{kumari2023ablating}, and SLD-M~\cite{schramowski2023safe} demonstrate limited effectiveness in erasing multiple concepts, which inadvertently results in their high specificity. UCE~\cite{gandikota2023unified} proves more effective, but its specificity decreases rapidly when more than 10 concepts are erased. Furthermore, it fails to maintain FID and CLIP score within a reasonable range when erasing more than 10 celebrities while preserving only 100. As to ESD-u~\cite{gandikota2023erasing} and ESD-x~\cite{gandikota2023erasing}, while effective, result in a lower proportion of facial images in their outputs (i.e., limited conceptual integrity), as shown in Figure~\ref{fig:cele_curve}~(f). This suggests that when their refined models are conditioned on erased celebrities, their outputs deviate from human likenesses, often leading to unpredictable and uncontrollable outcomes. This phenomenon is shown in Figure~\ref{fig:cele_erasure}, which presents a qualitative comparison of erasing 100 celebrities. In this comparison, Bill Clinton is in the erasure group, whereas Bill Murray and Amanda Seyfried are in the retention group. Notably, the preservation of Bill Murray's image poses a challenge due to his shared first name, `Bill,' with Bill Clinton in the erasure group. Our method effectively overcomes this issue.

\subsection{Explicit Content Erasure}
\label{sec:exp_explicit}
\noindent \textbf{Evaluation setup.} In this section, we attempt to mitigate the generation of explicit content in T2I models. We adopt the same setting used in SA \cite{heng2023selective}, finetuning SD v1.4 to erase four target phrases: `nudity', `naked', `erotic', and `sexual'. To evaluate efficacy and generality, we use each finetuned model to generate images using all 4,703 prompts sourced from the Inappropriate Image Prompt (I2P) dataset \cite{schramowski2023safe}. The NudeNet \cite{bedapudi2019nudenet} is employed to identify explicit content in these images, using a detection threshold of 0.6. Additionally, to assess specificity on regular content, we evaluate FID and CLIP score on the MS-COCO validation set, similar to the process described in Section \ref{sec:exp_cele}.

~\\
\noindent \textbf{Discussions and analysis.} Table~\ref{tab:nudity} presents our findings. Our refined model successfully generates the least amount of explicit content when conditioned on 4,703 prompts. Moreover, it showcases an impressive performance in FID, even surpassing the original SD v1.4. We note that such finetuning often does not have a consistent trend in improving or worsening FID and CLIP score on regular content generation. This pattern is also observed in the celebrity erasure, as shown in Figure~\ref{fig:cele_curve}~(d) and (e). Therefore, we consider the performance acceptable as long as FID and CLIP score remain within a reasonable range. It is also noteworthy that retraining SD v2.1 from scratch using a dataset curated to exclude explicit content yields only a minor improvement, compared with the original SD v1.4. Qualitative comparisons are provided in Appendix~\ref{sec:addtional_qualitative} for reference.

\begin{table*}[tbp]
	\caption{\textbf{Assessment of Explicit Content Removal:} (Left) Quantity of explicit content detected using the NudeNet detector on the I2P benchmark. (Right) Comparison of FID and CLIP on MS-COCO. The performance of the original SD~v1.4 is presented for reference. SD~v2.1 serves as a baseline that retrains the model from scratch on the curated dataset. $^\dag$: Results sourced from \cite{heng2023selective}. F: Female. M: Male.}
	\vspace{-0.6cm}
	\begin{center}
		\resizebox{1.0\textwidth}{!}{
			\begin{tabular}{lccccccccccc}
				\toprule
				\multirow{2}{*}{Method} & \multicolumn{9}{c}{Results of NudeNet Detection on I2P (Detected Quantity)} & \multicolumn{2}{c}{MS-COCO 30K} \\
				\cmidrule(lr){2-10}  \cmidrule(lr){11-12} 
				  & Armpits & Belly & Buttocks & Feet & Breasts (F) & Genitalia (F) & Breasts (M) & Genitalia (M) & {Total} $\downarrow$ & {FID} $\downarrow$ & {CLIP} $\uparrow$ \\
				\midrule
				FMN \cite{zhang2023forget} & 43 & 117 & 12 & 59  & 155 & 17 & 19 & 2 & 424 & 13.52 & 30.39 \\
				AC \cite{kumari2023ablating} & 153 & 180 & 45 & 66 & 298 & 22 & 67 & 7 & 838 & 14.13 & \textbf{31.37} \\
				UCE \cite{gandikota2023unified} & 29 & 62 & 7 & 29 & 35 & 5 & 11 & 4 & 182 & 14.07 & 30.85 \\
				SLD-M \cite{schramowski2023safe} & 47 & 72 & 3 & \textbf{21} & 39 & \textbf{1} & 26 & 3 & 212 & 16.34 & 30.90 \\
				ESD-x \cite{gandikota2023erasing} & 59 & 73 & 12 & 39 & 100 & 6 & 18 & 8 & 315 & 14.41 & 30.69 \\
				ESD-u \cite{gandikota2023erasing} & 32 & 30 & 2 & 19 & 27 & 3 & 8 & 2 & 123 & 15.10 & 30.21 \\
				SA$^\dag$ \cite{heng2023selective} & 72 & 77 & 19 & 25 & 83 & 16 & \textbf{0} & \textbf{0} & 292 & - & -\\
				Ours & \textbf{17} & \textbf{19} & \textbf{2} & 39 & \textbf{16} & 2 & 9 & 7 & \textbf{111} & \textbf{13.42} & 29.41 \\
				\midrule
				\rowcolor{white} 
				SD v1.4 \cite{rombach2022high} & 148 & 170 & 29 & 63 & 266 & 18 & 42 & 7 & 743 & 14.04 & 31.34 \\
				\rowcolor{white} 
				SD v2.1 \cite{rombach2022sd2} & 105 & 159 & 17 & 60 & 177 & 9 & 57 & 2 & 586 & 14.87 & 31.53 \\
				\bottomrule
		\end{tabular}}
	\end{center}
	\vspace{-0.3cm} 
	\label{tab:nudity}
\end{table*}

\begin{table}[tbp]
	\caption{\textbf{Assessment of Erasing 100 Artistic Styles:} $H_\text{a}$ indicates overall erasure performance.}
	\vspace{-0.6cm}
	\begin{center}
		\resizebox{0.48\textwidth}{!}{
			\begin{tabular}{lcc>{\columncolor{mypink}}ccc}
				\toprule
				Method & $\text{CLIP}_\text{e}$ $\downarrow$ & $\text{CLIP}_\text{s}$ $\uparrow$ & ${H_\text{a} }$ $\uparrow$ & FID-30K $\downarrow$ & CLIP-30K  $\uparrow$ \\
				\midrule
				FMN \cite{zhang2023forget} & 29.63 & \textbf{28.90} & -0.73 &  13.99 & \textbf{31.31}  \\
				AC \cite{kumari2023ablating} & 29.26 & 28.54 & -0.72 & 14.08  & 31.29 \\
				UCE \cite{gandikota2023unified} & 21.31 & 25.70 & 4.39 & 77.72 & 19.17   \\
				SLD-M \cite{schramowski2023safe} & 28.49 & 27.89 & -0.6 & 17.95  & 30.87  \\
				ESD-x \cite{gandikota2023erasing} & 20.89 & 21.21 & 0.32 & 15.19 & 29.52  \\
				ESD-u \cite{gandikota2023erasing} & \textbf{19.66} & 19.55 & -0.11 & 17.07 & 27.76  \\
				Ours & 22.59 & 28.58 & \textbf{5.99} & \textbf{12.71} &  29.51 \\
				\midrule
				\rowcolor{white} 
				SD v1.4 & 29.63 & 28.90 & - & 14.04 & 31.34  \\
				\bottomrule
		\end{tabular}}
	\end{center}
	\vspace{-0.5cm} 
	\label{tab:art}
\end{table}

\subsection{Artistic Style Erasure}
\label{sec:exp_art}
In this section, we evaluate our method and the baselines on erasing multiple artistic styles from SD v1.4. We utilize the Image Synthesis Style Studies Database \cite{art}, which compiles a list of artists whose styles can be replicated by SD v1.4. From this database, we sample 200 artists and split them into two groups: an erasure group of 100 artists and a retention group with 100 other artists. The complete list is provided in Table~\ref{tab:appendix_art}. To assess efficacy and specificity, we apply prompts like \textit{`Image in the style of \{artist name\}'} to both the erased and retained artists. We evaluate the erasure methods using two metrics: $\text{CLIP}_\text{e}$ and $\text{CLIP}_\text{s}$. The $\text{CLIP}_\text{e}$, which tests efficacy, is calculated between the prompts of the erased artists and the generated images. A lower $\text{CLIP}_\text{e}$ indicates better efficacy. Similarly, the $\text{CLIP}_\text{s}$, which assesses specificity, is calculated between the prompts of the retained artists and the generated images. A higher $\text{CLIP}_\text{s}$ signifies better specificity. 
We calculate the overall erasing capability by $H_\text{a} = \text{CLIP}_\text{s} - \text{CLIP}_\text{e}$. 
As reported in Table~\ref{tab:art}, our method also shows the superior ability to erase artistic styles on a large scale.

\subsection{Ablation Study}
\label{sec:exp_abl}
To study the impact of our key components, we conduct ablation studies on the challenging task of erasing 100 celebrities from SD v1.4. Different variations and their results are presented in Table~\ref{tab:ablation}. Variation 1 struggles to balance efficacy and specificity. When prioritizing prior preservation, its ability to erase is compromised. Variation 2, which trains LoRA without CFIS, restricts its specificity. Moreover, the naïve integration of LoRA exacerbates this issue, leading to poor specificity despite the successful erasure of the target concepts. Variation 3 fuses LoRA with closed-form fusion, which prevents interference from different LoRA modules, thereby improving specificity. However, without the CFIS, this configuration shows reduced training efficiency in erasure and decreased specificity.

\begin{table}[tbp]
	\caption{\textbf{Ablation Study on the Impact of Key Components in Erasing 100 Celebrities.} CFR: closed-form refinement. NLF: naïve LoRA fusion. CFLF: closed-form LoRA fusion. CFIS: concept-focal importance sampling. All presented values are denoted in percentage (\%).}
	\vspace{-0.6cm}
	\begin{center}
		\resizebox{0.48\textwidth}{!}{
			\begin{tabular}{lccccccc>{\columncolor{mypink}}c}
				\toprule
				\multirow{2}{*}{Config} & \multicolumn{5}{c}{Components} & \multicolumn{3}{c}{Metrics} \\
				\cmidrule(lr){2-6}  \cmidrule(lr){7-9} 
				 & CFR & LoRA & NLF & CFLF & CFIS & $\text{Acc}_\text{e}$ $\downarrow$ & $\text{Acc}_\text{s}$ $\uparrow$ & ${H_\text{c} }$ $\uparrow$  \\
				\midrule
				1 & \checkmark & $\times$ & $\times$ & $\times$ & $\times$ &67.79 & \textbf{85.05} & 46.72   \\
				2 & \checkmark & \checkmark & \checkmark & $\times$ & $\times$ &\textbf{0.08} & 32.16 & 48.66 \\
				3 & \checkmark & \checkmark & $\times$ & \checkmark & $\times$ & 18.70 & 61.78 & 70.21 \\
				\midrule
				Ours & \checkmark & \checkmark & $\times$ & \checkmark & \checkmark & 4.31 & 84.56 & \textbf{89.78}  \\
				\midrule
				SD v1.4 & - & - & - & - & - & 96.48 & 93.88 & 6.79  \\
				\bottomrule
		\end{tabular}}
	\end{center}
	\vspace{-0.2 cm} 
	\label{tab:ablation}
\end{table}

We also carry out ablation studies focused on independently adjusting either the `Key' or `Value' projection matrices, as detailed in Table~\ref{tab:ablation_key_value}. Intriguingly, exclusively finetuning the`Value' projection matrices for an identical number of steps can result in the deterioration of unrelated concepts, as indicated by a lower $\text{Acc}_\text{s}$. While fine-tuning only the `Value' projection matrices might seem efficient for obtaining satisfactory outcomes with minimal adjustments, its peak performance is notably inferior.

\begin{table}[tbp]
	\caption{\textbf{Ablation Study on the Impact of LoRA Finetuned Projection Matrices in Erasing 100 Celebrities.} All presented values are denoted in percentage (\%).}
	\vspace{-0.6cm}
	\begin{center}
		\resizebox{0.48\textwidth}{!}{
			\begin{tabular}{lllcc>{\columncolor{mypink}}c}
				\toprule
				\multirow{2}{*}{Config} & \multirow{2}{*}{Variation} & \multirow{2}{*}{Tuning Step} & \multicolumn{3}{c}{Metrics} \\
				 \cmidrule(lr){4-6} 
				& & & $\text{Acc}_\text{e}$ $\downarrow$ & $\text{Acc}_\text{s}$ $\uparrow$ & ${H_\text{c} }$ $\uparrow$  \\
				\midrule
				5 & Tune Key Only & 50 steps & 12.72 & 80.54 & 83.77   \\
				\midrule
				\multirow{6}{*}{6} & \multirow{6}{*}{Tune Value Only} & 50 steps & 0.77 & 38.65 & 55.63 \\
				 & & 25 steps & 3.28 & 62.74 & 76.11 \\
				 & & 10 steps & 10.47 & 77.81 & 83.26 \\
				 & & 5 steps & 12.72 & 80.45 & 83.73 \\
				 & & 3 steps & 14.46 & 81.70 & 83.58 \\
				 & & 1 steps & 15.39 & 82.37 & 83.48 \\
				\midrule
				Ours & Tune Key \& Value & 50 steps & 4.31 & 84.56 & \textbf{89.78}  \\
				\bottomrule
		\end{tabular}}
	\end{center}
	\vspace{-0.5 cm} 
	\label{tab:ablation_key_value}
\end{table}

\section{Additional Applications}

\begin{figure}[tbp]
	\centering
	\includegraphics[width=0.98\linewidth]{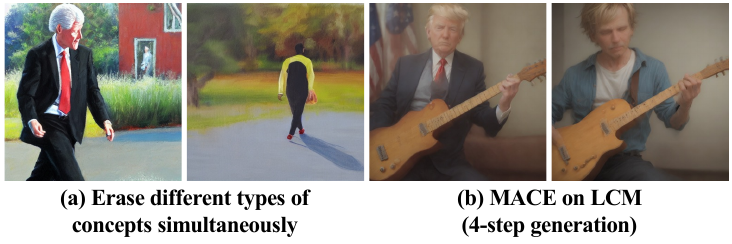}
	\vspace{-0.2cm}
	\caption{\textbf{(a) MACE can simultaneously erase different types of concepts.} The left image is generated from the original  SD~v1.4. The right one is generated by the MACE finetuned SD~v1.4 which erases the concepts of a celebrity (Bill Clinton) and a artistic style (Brent Heighton). Both images are generated using the prompt \textit{`Bill Clinton walking, Brent Heighton style'.} \textbf{(b) MACE is compatible with distilled diffusion models}. The left image is generated from the original LCM Dreamshaper v7. The right one is generated by the MACE finetuned LCM which erases the concept of `Trump'. Both images are generated using the prompt \textit{`a photo of Trump playing guitar'.}}
	\label{fig:add_app}
\end{figure}

MACE possesses the capability to simultaneously erase different types of concepts, such as both a celebrity likeness and artistic style, as shown in Figure~\ref{fig:add_app}~(a). Furthermore, MACE is compatible with distilled fast diffusion models (e.g., Latent Consistency Model \cite{luo2023latent}), with an example presented in Figure~\ref{fig:add_app}~(b).

\section{Limitations and Conclusion}
\label{sec:limitations}
Our proposed method, MACE, offers an effective solution for erasing mass concepts from T2I diffusion models. Our extensive experiments reveal that MACE achieves a remarkable balance between specificity and generality, particularly in erasing numerous concepts, surpassing the performance of prior methods. However, a discernible decline in the harmonic mean is observed as the number of erased concepts increases from 10 to 100. This trend could pose a limitation in erasing thousands of concepts from more advanced models in the future. Exploring ways to further scale up the erasure scope presents a crucial direction for future research. We believe MACE can be a pivotal tool for generative model service providers, empowering them to efficiently eliminate a variety of unwanted concepts. This is a vital step in releasing the next wave of advanced models, contributing to the creation of a safer AI community.
\section*{Acknowledgement}

This research is supported by the National Research Foundation, Singapore under its Strategic Capability Research Centres Funding Initiative. Any opinions, findings and conclusions or recommendations expressed in this material are those of the author(s) and do not reflect the views of National Research Foundation, Singapore.
{
    \small
    \bibliographystyle{ieeenat_fullname}
    \bibliography{paper_mace}
}

\clearpage
\onecolumn
\appendix

\noindent\textbf{\Large Appendix}
\renewcommand\thesection{\Alph{section}}
\setcounter{section}{0}




\section{Preliminaries}
\label{sec:preliminary}

\noindent \textbf{Latent diffusion model.} Our method is implemented using Stable Diffusion (SD), also known as Latent Diffusion Models (LDM)~\cite{rombach2022high}. This approach conducts the diffusion process within the latent space of an autoencoder. LDM is comprised of two principal components: a vector quantization autoencoder \cite{van2017neural,esser2021taming} and a diffusion model \cite{dhariwal2021diffusion, ho2020denoising, rombach2022high, sohl2015deep,kingma2021variational,song2020score}. The autoencoder undergoes pretraining to transform images into spatial latent codes via an encoder $\bfz = \mathcal{E}(\bfx)$, and it can reconstruct the images from these latent codes using a decoder $\bfx \approx \mathcal{D}(\mathcal{E}(\bfx))$. The diffusion model, on the other hand, is trained to generate latent codes that exist within the autoencoder's latent space. The training objective for the diffusion model is defined as follows \cite{ho2020denoising, rombach2022high}:
\begin{align} 
	\mathcal{L}_\text{LDM} = \mathbb{E}_{\bfz \sim \mathcal{E}(\bfx),\mathbf{c},\boldsymbol{\epsilon} \sim \mathcal{N}(0,1), t} \left[ \left\| \boldsymbol{\epsilon}  - \boldsymbol{\epsilon}_{\boldsymbol{\theta}}(\bfz_t, t, \mathbf{c}) \right\|^2_2 \right],
	\label{eq:appendix_ldm}
\end{align}
where $\bfz_t$ is the noisy latent, $t$ is the timestep, $\boldsymbol{\epsilon}$ is a standard Gaussian noise sample, $\boldsymbol{\epsilon}_{\boldsymbol{\theta}}$ is the denoising network, and $\mathbf{c}$ is the conditioning embeddings, which can be encoded from text prompts, class labels, segmentation masks, among others~\cite{rombach2022high}. During the inference process, Gaussian noise is sampled as a starting point $\bfz_T$ and successively denoised to produce a new latent code $\bfz_0$ through the well-trained denoising network $\boldsymbol{\epsilon}_{\boldsymbol{\theta}}$. Ultimately, this latent code is transformed into an image via the pretrained decoder $\bfx_0 \approx \mathcal{D}(\bfz_0)$.

~\\
\noindent \textbf{Cross-attention in text-to-image diffusion models.} In text-to-image diffusion models, cross-attention mechanisms serve as the pivotal interface for the interplay between image and text modalities. Initially, a text prompt undergoes tokenization, converting it into a series of unique token embeddings. These embeddings are then processed by a text encoder (e.g., CLIP \cite{radford2021learning} or T5 \cite{raffel2020exploring}), resulting in a final set of embeddings, denoted as $\boldsymbol{\mathcal{P}}=[\bfe_1~\bfe_2 \cdots \bfe_y ]$, wherein each token's embedding $\bfe_i$ is enriched with information from the entire token sequence. These enhanced embeddings are subsequently introduced into the cross-attention modules, where they act as navigational beacons for the image synthesis process. At certain layer $l$ and timestep $t$, the text embeddings are mapped using projection matrices, $\bfW_k$ and $\bfW_v$, to obtain the `Keys' $\bfk_{t,l} $ and `Values' $\bfv_{t,l} $, respectively. Concurrently, the image's features, $\bff_{t,l}$, undergo the projection $\bfW_q$ to form the `Queries' $\bfq_{t,l}$. The cross-attention mechanism computes the attention map as \cite{rombach2022high,vaswani2017attention}:
\begin{align} 
	\bfA_{t,l} = \text{softmax}\left( \frac{\bfq_{t,l} \cdot \bfk_{t,l}\tran }{\sqrt{d}} \right),
	\label{eq:appendix_attn}
\end{align}
where $d$ is the scaling factor to normalize the dot product. The module then synthesizes image features by aggregating `Values' with the attention weights, $\bfo_{t,l} = \bfA_{t,l} \cdot \bfv_{t,l}$. This process ensures that the generated images are intricately aligned with the input text, completing the text-to-image generation with high fidelity.

\section{Closed-Form Solution Proof}
\label{sec:proof}
In this section, we present a detailed derivation of the closed-form solution as written in Eq.~(\ref{eq:close-form-solution}). Our goal is to determine a refined matrix, denoted as $\bfW^{\prime}_k \in \mathbb{R}^{d_1 \times d_2}$, which encourages the model to refrain from embedding residual information of the target phrase into other words, while preserving the prior knowledge. The loss function is defined in Eq.~(\ref{eq:close-form-loss}), which is:
\begin{align*} 
	\mathcal{L}\left( \bfW^{\prime}_k \right) = \sum\limits_{i=1}^n \left\| \bfW_k^{\prime} \cdot \bfe^f_i - \bfW_k \cdot \bfe_i^g  \right\|_2^2 +  \lambda_1 \sum\limits_{i=n+1}^{n+m} \left\| \bfW_k^{\prime} \cdot \bfe^p_i - \bfW_k \cdot \bfe_i^p  \right\|_2^2,
\end{align*}
where $\lambda_1 \in \mathbb{R}^{+}$ is a hyperparameter, $\bfe_i^f \in \mathbb{R}^{d_2}$ is the embedding of a word co-existing with the target phrase, $\bfe_i^g \in \mathbb{R}^{d_2}$ is the embedding of that word when the target phrase is replaced with its super-category or a generic one, $\bfe_i^p \in \mathbb{R}^{d_2}$ is the embedding for preserving the prior, $\bfW_k \in \mathbb{R}^{d_1 \times d_2}$ is the pretrained weights, and $n,m$ are the number of embeddings for mapping and preserving, respectively. 

To seek the optimal $\bfW^{\prime}_k$, we differentiate the loss function with respect to it and set the derivative equal to zero:
\begin{align} 
	&\frac{\partial{\mathcal{L}\left( \bfW^{\prime}_k \right)}}{\partial \bfW^{\prime}_k} = 2\sum\limits_{i=1}^n \left( \bfW_k^{\prime} \cdot \bfe^f_i - \bfW_k \cdot \bfe_i^g  \right)(\bfe^f_i)\tran  +  2\lambda_1 \sum\limits_{i=n+1}^{n+m} \left(  \bfW_k^{\prime} \cdot \bfe^p_i - \bfW_k \cdot \bfe_i^p  \right) (\bfe^p_i)\tran = 0 \\
	&\sum\limits_{i=1}^n  \bfW_k^{\prime} \cdot \bfe^f_i \cdot (\bfe^f_i)\tran - \sum\limits_{i=1}^n  \bfW_k \cdot \bfe^g_i \cdot (\bfe^f_i)\tran + \lambda_1 \sum\limits_{i=n+1}^{n+m} \bfW_k^{\prime} \cdot \bfe^p_i \cdot (\bfe^p_i)\tran - \lambda_1 \sum\limits_{i=n+1}^{n+m} \bfW_k \cdot \bfe^p_i \cdot (\bfe^p_i)\tran = 0 \\
	&\sum\limits_{i=1}^n  \bfW_k^{\prime} \cdot \bfe^f_i \cdot (\bfe^f_i)\tran + \lambda_1 \sum\limits_{i=n+1}^{n+m} \bfW_k^{\prime} \cdot \bfe^p_i \cdot (\bfe^p_i)\tran =  \sum\limits_{i=1}^n  \bfW_k \cdot \bfe^g_i \cdot (\bfe^f_i)\tran + \lambda_1 \sum\limits_{i=n+1}^{n+m} \bfW_k \cdot \bfe^p_i \cdot (\bfe^p_i)\tran\\
	 &\bfW_k^{\prime} \left( \sum\limits_{i=1}^n \bfe^f_i \cdot (\bfe^f_i)\tran + \lambda_1 \sum\limits_{i=n+1}^{n+m} \bfe^p_i \cdot (\bfe^p_i)\tran \right) =  \sum\limits_{i=1}^n  \bfW_k \cdot \bfe^g_i \cdot (\bfe^f_i)\tran + \lambda_1 \sum\limits_{i=n+1}^{n+m} \bfW_k \cdot \bfe^p_i \cdot (\bfe^p_i)\tran \\
	 &\bfW_k^{\prime} =  \left(\sum\limits_{i=1}^n  \bfW_k \cdot \bfe^g_i \cdot (\bfe^f_i)\tran + \lambda_1 \sum\limits_{i=n+1}^{n+m} \bfW_k \cdot \bfe^p_i \cdot (\bfe^p_i)\tran \right) \cdot \left( \sum\limits_{i=1}^n \bfe^f_i \cdot (\bfe^f_i)\tran + \lambda_1 \sum\limits_{i=n+1}^{n+m} \bfe^p_i \cdot (\bfe^p_i)\tran \right)^{-1}.
\end{align}
To ensure the validity of the final step, it is crucial that the symmetric real matrix 
$\left( \sum_{i=1}^n \bfe^f_i \cdot (\bfe^f_i)\tran + \lambda_1 \sum_{i=n+1}^{n+m} \bfe^p_i \cdot (\bfe^p_i)\tran \right)$
is full rank. 
For any non-zero vector $\bfx \in \mathbb{R}^{d_2}$, we examine the following quadratic form:
\begin{align} 
	\bfx\tran \cdot \left( \sum_{i=1}^n \bfe^f_i \cdot (\bfe^f_i)\tran + \lambda_1 \sum_{i=n+1}^{n+m} \bfe^p_i \cdot (\bfe^p_i)\tran \right) \cdot \bfx &= \sum_{i=1}^{n} \left( \bfx\tran \bfe^f_i \right) \cdot \left( \bfx\tran \bfe^f_i \right)\tran + \lambda_1 \sum_{i=n+1}^{n+m} \left( \bfx\tran \bfe^p_i \right) \cdot \left( \bfx\tran \bfe^p_i \right)\tran \\
	&= \sum_{i=1}^{n} \left\| \bfx\tran \bfe^f_i \right\|^2_2 + \lambda_1 \sum_{i=n+1}^{n+m} \left\| \bfx\tran \bfe^p_i \right\|^2_2 \ge 0.
\end{align}
The prior preserving embeddings $\bfe_i^p$ are computed by default using the MS-COCO dataset \cite{lin2014microsoft}. Due to the extensive number of terms involved in the summation, it is highly improbable for all terms $\left\| \bfx\tran \bfe^p_i \right\|^2_2$ in the sum to equal zero. Hence, in general cases, the matrix $\left( \sum_{i=1}^n \bfe^f_i \cdot (\bfe^f_i)\tran + \lambda_1 \sum_{i=n+1}^{n+m} \bfe^p_i \cdot (\bfe^p_i)\tran \right)$ is positive definite and thus invertible. 
The derivation is applicable to $\bfW_v^{\prime}$ as well. 

In addition to retaining general prior knowledge, akin to UCE~\cite{gandikota2023unified}, our framework extends support to allow users to highlight and preserve domain-specific concepts. This functionality is absent in most preceding frameworks. For instance, when two concepts share a strong correlation, removing one could potentially impair the generation quality of the other, which might be intended for preservation. Both general and domain-specific prior knowledge can be incorporated into the second term of Eq.~(\ref{eq:close-form-loss}). We set a weighting factor $\lambda_3$ to calibrate the significance attributed to each type of knowledge. Thus, Eq.~(\ref{eq:close-form-solution}) can be reformulated as follows:
\begin{equation} 
	\label{eq:close-form-fuse}
	\begin{split}
		\bfW_k^{\prime} =  & \left(\sum\limits_{i=1}^n  \bfW_k  \cdot \bfe^g_i  \cdot (\bfe^f_i)\tran + \lambda_1 \sum\limits_{i=n+1}^{n+m^\prime} \bfW_k  \cdot \bfe^p_i \cdot (\bfe^p_i)\tran + \lambda_3 \sum\limits_{i=n+m^\prime}^{n+m} \bfW_k \cdot \bfe^p_i \cdot (\bfe^p_i)\tran \right) \\
		\cdot & \left( \sum\limits_{i=1}^n \bfe^f_i \cdot (\bfe^f_i)\tran + \lambda_1 \sum\limits_{i=n+1}^{n+m^\prime} \bfe^p_i \cdot (\bfe^p_i)\tran + \lambda_3 \sum\limits_{i=n+m^\prime}^{n+m} \bfe^p_i \cdot (\bfe^p_i)\tran \right)^{-1},
	\end{split}
\end{equation}
where we have $m^\prime$ terms of general knowledge and $m-m^\prime$ terms of domain-specific knowledge.

\section{Implementation Details}
\label{sec:imple_details}
\subsection{Experimental Setup Details}
\begin{table*}[tb]
	\caption{The synonyms and mapping concepts for the ten object classes in the CIFAR-10 dataset.}
	\vspace{-0.6cm}
	\begin{center}
		\resizebox{1\textwidth}{!}{
			\begin{tabular}{p{3.0cm} | p{1.5cm} | p{1.5cm} | p{1.5cm} | p{1.5cm} | p{1.5cm} | p{1.5cm} | p{1.5cm} | p{1.5cm} | p{1.5cm} | p{1.5cm} }
				\hline
				\makecell[l]{\textbf{Object Classes}} & Airplane & Automobile & Bird & Cat & Deer & Dog & Frog & Horse & Ship & Truck \\
				\hline
				
				\multirow{6}{*}{\textbf{Synonyms}} 
				& \multirow{2}{*}{Aircraft} & \multirow{2}{*}{Car} & \multirow{2}{*}{Avian} & \multirow{2}{*}{Feline} & \multirow{2}{*}{Hart} & \multirow{2}{*}{Canine} & \multirow{2}{*}{Amphibian} & \multirow{2}{*}{Equine} & \multirow{2}{*}{Vessel} & \multirow{2}{*}{Lorry} \\
				&&&&&&&&&& \\
				& \multirow{2}{*}{Plane} & \multirow{2}{*}{Vehicle} & \multirow{2}{*}{Fowl} & \multirow{2}{*}{Kitty} & \multirow{2}{*}{Stag} & \multirow{2}{*}{Pooch} & \multirow{2}{*}{Anuran} & \multirow{2}{*}{Steed} & \multirow{2}{*}{Boat} & \multirow{2}{*}{Rig} \\
				&&&&&&&&&& \\
				& Jet & Motorcar & Winged Creature  & Housecat & Doe & Hound & Tadpole & Mount & Watercraft & Hauler\\
				\hline
				\multirow{2}{*}{\makecell[l]{\textbf{Mapping Concepts} \\ (Randomly Sampled)}} & \multirow{2}{*}{Ground} & \multirow{2}{*}{Sky} & \multirow{2}{*}{Street}  & \multirow{2}{*}{Forest} & \multirow{2}{*}{Street} & \multirow{2}{*}{Sky} & \multirow{2}{*}{Forest} & \multirow{2}{*}{Forest} & \multirow{2}{*}{Ground} & \multirow{2}{*}{Sky} \\
				& & & & & & & & & & \\
				\hline
		\end{tabular}}
	\end{center}
	\vspace{-0.5cm} 
	\label{tab:appendix_syn}
\end{table*}

\noindent \textbf{Object erasure.} To assess the generality of erasure, we prepare three synonyms for each of the ten object classes in the CIFAR-10 dataset \cite{krizhevsky2009learning}. These synonyms are listed in Table~\ref{tab:appendix_syn}. Since the object classes lack proper super-categories, we allocate generic concepts (e.g., sky or ground) to them, as also presented in Table~\ref{tab:appendix_syn}. To evaluate the erasure capability of different methods, we use each finetuned model to generate 200, 600, and 1,800 images for three categories: the erased object (efficacy), its three synonyms (generality), and nine remaining objects (specificity). For the erased object, the prompt is \textit{`a photo of the \{erased class name\}'}. For the three synonyms of the erased object, we use the prompt \textit{`a photo of the \{synonym of erased class name\}'}. Lastly, for each of the nine remaining unaltered objects, the prompt is \textit{`a photo of the \{unaltered class name\}'}.

~\\
\noindent \textbf{Celebrity erasure.} For assessing the task of multiple celebrity erasure, we establish a dataset consisting of 200 celebrities. Their portraits generated by SD v1.4 are recognizable with exceptional accuracy by the GIPHY Celebrity Detector (GCD)~\cite{GCD}. This dataset is divided into two groups: an erasure group with 100 celebrities and a retention group with 100 other celebrities. The full list for both groups is presented in Table~\ref{tab:appendix_cele}. For evaluating the task of celebrity erasure, we use five text prompts to generate images. These prompts are as follows: \textit{`a portrait of \{celebrity name\}’, `a sketch of \{celebrity name\}’, `an oil painting of \{celebrity name\}’, `\{celebrity name\} in an official photo’,} and \textit{`an image capturing \{celebrity name\} at a public event’.} The detailed breakdown of the number of images generated for each experiment is presented in Table~\ref{tab:appendix_cele1}.

\begin{table*}[tbp]
	\vspace{-0.2cm} 
	\caption{\textbf{The Evaluation Setup for Celebrity Erasure:} Our celebrity dataset contains an erasure group with 100 celebrities and a retention group with another 100 celebrities. Portraits of these celebrities can be effectively generated using SD~v1.4. The generated portraits are accurately recognizable by the GIPHY Celebrity Detector (GCD) with an accuracy exceeding 99\%. To perform erasure experiments involving 1, 5, 10, and 100 celebrities, a corresponding number of celebrities are selected from the erasure group for each experiment. In all cases, the entire retention group is utilized.}
	\vspace{-0.5cm}
	\begin{center}
		\resizebox{1\textwidth}{!}{
			\begin{tabular}{ p{1.4cm}<{\centering} | p{2.4cm}<{\centering} |  p{1.6cm}<{\centering} | p{14 cm}}
				\hline
				Group & \makecell[l]{\# of Celebrities \\ to Be Erased} & \makecell[c]{Mapping \\ Concept} & Celebrity \\
				\hline
				\multirow{22}{*}{\makecell{Erasure \\ Group}} & 1 & `a woman' & \textit{`Melania Trump'} \\
				\cline{2-4}
				& 5 & `a person' & \textit{`Adam Driver', `Adriana Lima', `Amber Heard', `Amy Adams', `Andrew Garfield'} \\
				\cline{2-4}
				& \multirow{2}{*}{10} & \multirow{2}{*}{`a person'} & \textit{`Adam Driver', `Adriana Lima', `Amber Heard', `Amy Adams', `Andrew Garfield', `Angelina Jolie', `Anjelica Huston', `Anna Faris', `Anna Kendrick', `Anne Hathaway'} \\
				\cline{2-4}
				& \multirow{18}{*}{100} & \multirow{18}{*}{`a person'} & \textit{`Adam Driver', `Adriana Lima', `Amber Heard', `Amy Adams', `Andrew Garfield', `Angelina Jolie', `Anjelica Huston', `Anna Faris', `Anna Kendrick', `Anne Hathaway', `Arnold Schwarzenegger', `Barack Obama', `Beth Behrs', `Bill Clinton', `Bob Dylan', `Bob Marley', `Bradley Cooper', `Bruce Willis', `Bryan Cranston', `Cameron Diaz', `Channing Tatum', `Charlie Sheen', `Charlize Theron', `Chris Evans', `Chris Hemsworth', `Chris Pine', `Chuck Norris', `Courteney Cox', `Demi Lovato', `Drake', `Drew Barrymore', `Dwayne Johnson', `Ed Sheeran', `Elon Musk', `Elvis Presley', `Emma Stone', `Frida Kahlo', `George Clooney', `Glenn Close', `Gwyneth Paltrow', `Harrison Ford', `Hillary Clinton', `Hugh Jackman', `Idris Elba', `Jake Gyllenhaal', `James Franco', `Jared Leto', `Jason Momoa', `Jennifer Aniston', `Jennifer Lawrence', `Jennifer Lopez', `Jeremy Renner', `Jessica Biel', `Jessica Chastain', `John Oliver', `John Wayne', `Johnny Depp', `Julianne Hough', `Justin Timberlake', `Kate Bosworth', `Kate Winslet', `Leonardo Dicaprio', `Margot Robbie', `Mariah Carey', `Melania Trump', `Meryl Streep', `Mick Jagger', `Mila Kunis', `Milla Jovovich', `Morgan Freeman', `Nick Jonas', `Nicolas Cage', `Nicole Kidman', `Octavia Spencer', `Olivia Wilde', `Oprah Winfrey', `Paul Mccartney', `Paul Walker', `Peter Dinklage', `Philip Seymour Hoffman', `Reese Witherspoon', `Richard Gere', `Ricky Gervais', `Rihanna', `Robin Williams', `Ronald Reagan', `Ryan Gosling', `Ryan Reynolds', `Shia Labeouf', `Shirley Temple', `Spike Lee', `Stan Lee', `Theresa May', `Tom Cruise', `Tom Hanks', `Tom Hardy', `Tom Hiddleston', `Whoopi Goldberg', `Zac Efron', `Zayn Malik'} \\
				\hline
				\multirow{18}{*}{\makecell{Retention \\ Group}} & \multirow{18}{*}{\makecell{1, 5, 10, and 100}} & \multirow{18}{*}{\makecell{-}} & \textit{`Aaron Paul', `Alec Baldwin', `Amanda Seyfried', `Amy Poehler', `Amy Schumer', `Amy Winehouse', `Andy Samberg', `Aretha Franklin', `Avril Lavigne', `Aziz Ansari', `Barry Manilow', `Ben Affleck', `Ben Stiller', `Benicio Del Toro', `Bette Midler', `Betty White', `Bill Murray', `Bill Nye', `Britney Spears', `Brittany Snow', `Bruce Lee', `Burt Reynolds', `Charles Manson', `Christie Brinkley', `Christina Hendricks', `Clint Eastwood', `Countess Vaughn', `Dakota Johnson', `Dane Dehaan', `David Bowie', `David Tennant', `Denise Richards', `Doris Day', `Dr Dre', `Elizabeth Taylor', `Emma Roberts', `Fred Rogers', `Gal Gadot', `George Bush', `George Takei', `Gillian Anderson', `Gordon Ramsey', `Halle Berry', `Harry Dean Stanton', `Harry Styles', `Hayley Atwell', `Heath Ledger', `Henry Cavill', `Jackie Chan', `Jada Pinkett Smith', `James Garner', `Jason Statham', `Jeff Bridges', `Jennifer Connelly', `Jensen Ackles', `Jim Morrison', `Jimmy Carter', `Joan Rivers', `John Lennon', `Johnny Cash', `Jon Hamm', `Judy Garland', `Julianne Moore', `Justin Bieber', `Kaley Cuoco', `Kate Upton', `Keanu Reeves', `Kim Jong Un', `Kirsten Dunst', `Kristen Stewart', `Krysten Ritter', `Lana Del Rey', `Leslie Jones', `Lily Collins', `Lindsay Lohan', `Liv Tyler', `Lizzy Caplan', `Maggie Gyllenhaal', `Matt Damon', `Matt Smith', `Matthew Mcconaughey', `Maya Angelou', `Megan Fox', `Mel Gibson', `Melanie Griffith', `Michael Cera', `Michael Ealy', `Natalie Portman', `Neil Degrasse Tyson', `Niall Horan', `Patrick Stewart', `Paul Rudd', `Paul Wesley', `Pierce Brosnan', `Prince', `Queen Elizabeth', `Rachel Dratch', `Rachel Mcadams', `Reba Mcentire', `Robert De Niro'} \\
				\hline
		\end{tabular}}
	\end{center}
	\label{tab:appendix_cele}
\end{table*}

\begin{table*}[tbp]
	\caption{The detailed breakdown of the number (\#) of images generated for each celebrity erasure experiment.}
	\vspace{-0.6cm}
	\begin{center}
		\resizebox{0.7\textwidth}{!}{
			\begin{tabular}{m{2.5cm}<{\centering} | m{3cm}<{\centering} | m{4cm}<{\centering} | m{3cm}<{\centering} }
				\hline
				\makecell{\# of Celebrities \\ to Be Erased} & {Celebrity Group} & \makecell{\# of Images Generated \\ for Each Celebrity} & \makecell{Total \# of \\ Generated Images} \\
				\hline
				\multirow{2}{*}{1} & Erasure Group & 250 & 250 \\
				\cline{2-4}
				& Retention Group & 25 & 2500 \\
				
				\hline
				\multirow{2}{*}{5} & Erasure Group & 50 & 250 \\
				\cline{2-4}
				& Retention Group & 25 & 2500 \\
				
				\hline
				\multirow{2}{*}{10} & Erasure Group & 25 & 250 \\
				\cline{2-4}
				& Retention Group & 25 & 2500 \\
				
				\hline
				\multirow{2}{*}{100} & Erasure Group & 25 & 2500 \\
				\cline{2-4}
				& Retention Group & 25 & 2500 \\
				
				\hline
		\end{tabular}}
	\end{center}
	\vspace{-0.3cm} 
	\label{tab:appendix_cele1}
\end{table*}

~\\
\noindent \textbf{Explicit content erasure.} We adopt the same setting used in SA \cite{heng2023selective} to erase `nudity’, `naked’, `erotic’, and `sextual’ from SD~v1.4. The mapping concept is set as `a person wearing clothes’.

~\\
\noindent \textbf{Artistic style erasure.} We utilize the Image Synthesis Style Studies Database \cite{art}, which compiles a list of artists whose styles can be replicated by SD v1.4. From this database, we sample 200 artists and split them into two groups: an erasure group of 100 artists and a retention group with 100 other artists. The full list for both groups is presented in Table~\ref{tab:appendix_art}. To assess efficacy and specificity, we apply the same five prompts and seeds as in \cite{gandikota2023unified} for both the erased and retained artists group. These prompts include \textit{`Image in the style of \{artist name\}', `Art inspired by \{artist name\}', `Painting in the style of \{artist name\}', `A reproduction of art by \{artist name\}'} and \textit{`A famous artwork by \{artist name\}'}. For each of 100 artists, we use each prompt to generate five images, resulting in 25 images per artist. Thus, this yields 2500 images for each group.

\begin{table*}[tbp]
	\caption{\textbf{The Evaluation Setup for Artistic Style Erasure:} We sample 200 artists from the Image Synthesis Style Studies Database \cite{art}. They are split into two groups: an erasure group with 100 artists and a retention group with another 100 artists. The artworks of these artists can be successfully replicated by SD~v1.4.}
	\vspace{-0.5cm}
	\begin{center}
		\resizebox{1\textwidth}{!}{
			\begin{tabular}{ p{1.4cm}<{\centering} | p{2.65 cm}<{\centering} |  p{1.6cm}<{\centering} | p{14 cm}}
				\hline
				Group & \makecell[l]{\# of Artistic Styles \\ to Be Erased} & \makecell[c]{Mapping \\ Concept} & Artist \\
				\hline
				\multirow{18}{*}{\makecell{Erasure \\ Group}} & \multirow{18}{*}{100} & \multirow{18}{*}{`art'} & \textit{`Brent Heighton', `Brett Weston', `Brett Whiteley', `Brian Bolland', `Brian Despain', `Brian Froud', `Brian K. Vaughan', `Brian Kesinger', `Brian Mashburn', `Brian Oldham', `Brian Stelfreeze', `Brian Sum', `Briana Mora', `Brice Marden', `Bridget Bate Tichenor', `Briton Rivière', `Brooke Didonato', `Brooke Shaden', `Brothers Grimm', `Brothers Hildebrandt', `Bruce Munro', `Bruce Nauman', `Bruce Pennington', `Bruce Timm', `Bruno Catalano', `Bruno Munari', `Bruno Walpoth', `Bryan Hitch', `Butcher Billy', `C. R. W. Nevinson', `Cagnaccio Di San Pietro', `Camille Corot', `Camille Pissarro', `Camille Walala', `Canaletto', `Candido Portinari', `Carel Willink', `Carl Barks', `Carl Gustav Carus', `Carl Holsoe', `Carl Larsson', `Carl Spitzweg', `Carlo Crivelli', `Carlos Schwabe', `Carmen Saldana', `Carne Griffiths', `Casey Weldon', `Caspar David Friedrich', `Cassius Marcellus Coolidge', `Catrin Welz-Stein', `Cedric Peyravernay', `Chad Knight', `Chantal Joffe', `Charles Addams', `Charles Angrand', `Charles Blackman', `Charles Camoin', `Charles Dana Gibson', `Charles E. Burchfield', `Charles Gwathmey', `Charles Le Brun', `Charles Liu', `Charles Schridde', `Charles Schulz', `Charles Spencelayh', `Charles Vess', `Charles-Francois Daubigny', `Charlie Bowater', `Charline Von Heyl', `Chaïm Soutine', `Chen Zhen', `Chesley Bonestell', `Chiharu Shiota', `Ching Yeh', `Chip Zdarsky', `Chris Claremont', `Chris Cunningham', `Chris Foss', `Chris Leib', `Chris Moore', `Chris Ofili', `Chris Saunders', `Chris Turnham', `Chris Uminga', `Chris Van Allsburg', `Chris Ware', `Christian Dimitrov', `Christian Grajewski', `Christophe Vacher', `Christopher Balaskas', `Christopher Jin Baron', `Chuck Close', `Cicely Mary Barker', `Cindy Sherman', `Clara Miller Burd', `Clara Peeters', `Clarence Holbrook Carter', `Claude Cahun', `Claude Monet', `Clemens Ascher'} \\
				\hline
				\multirow{18}{*}{\makecell{Retention \\ Group}} & \multirow{18}{*}{\makecell{100}} & \multirow{18}{*}{\makecell{-}} & \textit{`A.J.Casson', `Aaron Douglas', `Aaron Horkey', `Aaron Jasinski', `Aaron Siskind', `Abbott Fuller Graves', `Abbott Handerson Thayer', `Abdel Hadi Al Gazzar', `Abed Abdi', `Abigail Larson', `Abraham Mintchine', `Abraham Pether', `Abram Efimovich Arkhipov', `Adam Elsheimer', `Adam Hughes', `Adam Martinakis', `Adam Paquette', `Adi Granov', `Adolf Hirémy-Hirschl', `Adolph Gottlieb', `Adolph Menzel', `Adonna Khare', `Adriaen van Ostade', `Adriaen van Outrecht', `Adrian Donoghue', `Adrian Ghenie', `Adrian Paul Allinson', `Adrian Smith', `Adrian Tomine', `Adrianus Eversen', `Afarin Sajedi', `Affandi', `Aggi Erguna', `Agnes Cecile', `Agnes Lawrence Pelton', `Agnes Martin', `Agostino Arrivabene', `Agostino Tassi', `Ai Weiwei', `Ai Yazawa', `Akihiko Yoshida', `Akira Toriyama', `Akos Major', `Akseli Gallen-Kallela', `Al Capp', `Al Feldstein', `Al Williamson', `Alain Laboile', `Alan Bean', `Alan Davis', `Alan Kenny', `Alan Lee', `Alan Moore', `Alan Parry', `Alan Schaller', `Alasdair McLellan', `Alastair Magnaldo', `Alayna Lemmer', `Albert Benois', `Albert Bierstadt', `Albert Bloch', `Albert Dubois-Pillet', `Albert Eckhout', `Albert Edelfelt', `Albert Gleizes', `Albert Goodwin', `Albert Joseph Moore', `Albert Koetsier', `Albert Kotin', `Albert Lynch', `Albert Marquet', `Albert Pinkham Ryder', `Albert Robida', `Albert Servaes', `Albert Tucker', `Albert Watson', `Alberto Biasi', `Alberto Burri', `Alberto Giacometti', `Alberto Magnelli', `Alberto Seveso', `Alberto Sughi', `Alberto Vargas', `Albrecht Anker', `Albrecht Durer', `Alec Soth', `Alejandro Burdisio', `Alejandro Jodorowsky', `Aleksey Savrasov', `Aleksi Briclot', `Alena Aenami', `Alessandro Allori', `Alessandro Barbucci', `Alessandro Gottardo', `Alessio Albi', `Alex Alemany', `Alex Andreev', `Alex Colville', `Alex Figini', `Alex Garant'} \\
				\hline
		\end{tabular}}
	\end{center}
		\vspace{-0.8cm} 
	\label{tab:appendix_art}
\end{table*}

\subsection{Training Configurations}

\noindent \textbf{Implementation of previous works.} In our series of four experiments, we focus on comparing our proposed method with existing methods, including ESD-u\footnote{\scriptsize \url{https://github.com/rohitgandikota/erasing}} \cite{gandikota2023erasing}, ESD-x \cite{gandikota2023erasing}, FMN\footnote{\scriptsize \url{https://github.com/SHI-Labs/Forget-Me-Not}} \cite{zhang2023forget}, SLD-M\footnote{\scriptsize \url{https://github.com/ml-research/safe-latent-diffusion}} \cite{schramowski2023safe}, UCE\footnote{\scriptsize \url{https://github.com/rohitgandikota/unified-concept-editing}} \cite{gandikota2023unified}, AC\footnote{\scriptsize \url{https://github.com/nupurkmr9/concept-ablation}} \cite{kumari2023ablating}, and SA\footnote{\scriptsize \url{https://github.com/clear-nus/selective-amnesia}} \cite{heng2023selective}. Notably, SA~\cite{heng2023selective} demands extensive resources, requiring 4 RTX A6000s and over 12 hours of training for concept erasure. Consequently, we have not replicated their findings due to these extensive requirements. Instead, we align our explicit content erasure task with SA's settings \cite{heng2023selective}, and we employ their reported experimental results for our comparative analysis. Beyond SA \cite{heng2023selective}, we implement each existing method following their recommended configurations for various erasure types (such as objects, style, or nudity). It is important to note that several methods (e.g., FMN \cite{zhang2023forget} and AC  \cite{kumari2023ablating}) are not tailored for erasing multiple concepts. In our preliminary tests, we observe that without altering the algorithm or further tuning the suggested parameters, training for multiple concepts—either sequentially or in parallel—yielded comparably mediocre results, marked by either inadequate specificity or generality. Consequently, we opt for a parallel training manner for them when erasing multiple concepts to save resources.

~\\
\noindent \textbf{Implementation of MACE.} This section details the implementation of MACE, focusing on the hyperparameters applied across various experimental scenarios, as outlined in Table~\ref{tab:appendix_training_config}. For the erasure of explicit content, we leverage general prior knowledge estimated from the MSCOCO dataset, without incorporating any domain-specific knowledge. For the erasure of celebrity likenesses and artistic styles, MACE again utilizes the general prior knowledge from the MSCOCO dataset. Additionally, domain-specific knowledge is employed, which is calculated based on the corresponding retention groups that users wish to preserve. Object erasure presents a special case where the prior knowledge from the MSCOCO dataset includes the concepts we aim to erase (e.g., cat, dog, or airplane). Thus, a direct application of the standard approach is not feasible. To address this, we modify the loss function. Instead of using the second term in Eq.~(\ref{eq:close-form-loss}), we use $\left\| \bfW_k^\prime - \bfW_k \right\|^2_2 $ to preserve the original knowledge.


\begin{table*}[htbp]
	\caption{\textbf{Hyperparameters Utilized in MACE Across Different Experimental Sets.}}
	\vspace{-0.3cm}
	\begin{center}
		\resizebox{1.0\textwidth}{!}{
			\begin{tabular}{m{3cm}<{\centering} | m{3cm}<{\centering} | m{3cm}<{\centering} | m{2cm}<{\centering} | m{2cm}<{\centering} | m{2cm}<{\centering} | m{2cm}<{\centering}}
				\hline
				Erasure Type & Segment & LoRA Training Step & Learning Rate & $\lambda_1 = \lambda_2$ & $\lambda_3$ & Rank $r$ \\
				\hline
				\multirow{10}{*}{\makecell[c]{Object}} & Airplane & 50 & $1.0 \times 10^{-5}$ & $1000.0$ & - & 1 \\
				& Automobile & 50 & $1.0 \times 10^{-5}$ & $100.0$ & - & 1 \\
				& Bird & 50 & $1.0 \times 10^{-5}$ & $10.0$ & - & 1 \\
				& Cat & 50 & $1.0 \times 10^{-5}$ & $1000.0$ & - & 1 \\
				& Deer & 50 & $1.0 \times 10^{-5}$ & $10.0$ & - & 1 \\
				& Dog & 50 & $1.0 \times 10^{-5}$ & $10.0$ & - & 1 \\
				& Frog & 50 & $1.0 \times 10^{-5}$ & $0.4$ & - & 1 \\
				& Horse & 50 & $1.0 \times 10^{-5}$ & $1.0$ & - & 1 \\
				& Ship & 50 & $1.0 \times 10^{-5}$ & $1000.0$ & - & 1 \\
				& Truck & 50 & $1.0 \times 10^{-5}$ & $0.1$ & - & 1 \\
				\hline
				\multirow{4}{*}{\makecell[c]{Celebrity}} 
				& 1 Celebrity & 50 & $1.0 \times 10^{-4}$ & $1.0 \times 10^{-4}$ & $0.8$ & 1 \\
				& 5 Celebrities & 50 & $1.0 \times 10^{-4}$ & $1.0 \times 10^{-4}$ & $5.0$ & 1 \\
				& 10 Celebrities & 50 & $1.0 \times 10^{-4}$ & $1.0 \times 10^{-4}$ & $8.0$ & 1 \\
				& 100 Celebrities & 50 & $1.0 \times 10^{-4}$ & $1.0 \times 10^{-4}$ & $20.0$ & 1 \\
				\hline
				{Artistic Style}
				& 100 Artistic Styles & 50 & $1.0 \times 10^{-4}$ & $1.0 \times 10^{-4}$ & $8.0$ & 1 \\
				\hline
				{Explicit Content} & {‘Nudity’, ‘Naked’, ‘Erotic’, ‘Sexual’.} & 120 & $1.0 \times 10^{-5}$ & $7.0 \times 10^{-7}$ & - & 1 \\
				\hline
		\end{tabular}}
	\end{center}
	\label{tab:appendix_training_config}
\end{table*}

\section{Additional Evaluation Results of Erasing the CIFAR-10 Classes}
\label{sec:additional_cifar_10}

Table~\ref{tab:appendix_object} presents the results of erasing the final six object classes of the CIFAR-10 dataset \cite{krizhevsky2009learning}. Our approach achieves the highest harmonic mean across the erasure of these six object classes. This highlights the exceptional erasure capabilities of our approach, effectively balancing specificity and generality. 

\begin{table*}[htbp]
	\caption{\textbf{Evaluation of Erasing the CIFAR-10 Classes:} Results for the final six individual classes are presented. CLIP classification accuracies are reported for each erased class in three sets: the erased class itself ($\text{Acc}_\text{e}$, efficacy), the nine remaining unaffected classes ($\text{Acc}_\text{s}$, specificity), and three synonyms of the erased class ($\text{Acc}_\text{g}$, generality). The \textcolor{mypink1}{harmonic means} $H_\text{o}$ reflect the comprehensive erasure capability. All presented values are denoted in percentage (\%). The classification accuracies of images generated by the original SD v1.4 are presented for reference.}
	\vspace{-0.3cm}
	\begin{center}
		\resizebox{1\textwidth}{!}{
			\begin{tabular}{lccc>{\columncolor{mypink}}cccc>{\columncolor{mypink}}cccc>{\columncolor{mypink}}cccc>{\columncolor{mypink}}cccc>{\columncolor{mypink}}cccc>{\columncolor{mypink}}c}
				\toprule
				\multirow{2}{*}{Method} & \multicolumn{4}{c}{Deer Erased} & \multicolumn{4}{c}{Dog Erased} & \multicolumn{4}{c}{Frog Erased} & \multicolumn{4}{c}{Horse Erased} & \multicolumn{4}{c}{Ship Erased} & \multicolumn{4}{c}{Truck Erased} \\
				\cmidrule(lr){2-5}  \cmidrule(lr){6-9} \cmidrule(lr){10-13} \cmidrule(lr){14-17} \cmidrule(lr){18-21}  \cmidrule(lr){22-25}  
				& $\text{Acc}_\text{e}$ $\downarrow$ & $\text{Acc}_\text{s}$ $\uparrow$ & $\text{Acc}_\text{g}$ $\downarrow$ & ${H_\text{o} }$ $\uparrow$ & $\text{Acc}_\text{e}$ $\downarrow$ & $\text{Acc}_\text{s}$ $\uparrow$ & $\text{Acc}_\text{g}$ $\downarrow$ & ${H_\text{o} }$ $\uparrow$ & $\text{Acc}_\text{e}$ $\downarrow$ & $\text{Acc}_\text{s}$ $\uparrow$ & $\text{Acc}_\text{g}$ $\downarrow$ & ${H_\text{o} }$ $\uparrow$ & $\text{Acc}_\text{e}$ $\downarrow$ & $\text{Acc}_\text{s}$ $\uparrow$ & $\text{Acc}_\text{g}$ $\downarrow$ & ${H_\text{o} }$ $\uparrow$ & $\text{Acc}_\text{e}$ $\downarrow$ & $\text{Acc}_\text{s}$ $\uparrow$ & $\text{Acc}_\text{g}$ $\downarrow$ & ${H_\text{o} }$ $\uparrow$ & $\text{Acc}_\text{e}$ $\downarrow$ & $\text{Acc}_\text{s}$ $\uparrow$ & $\text{Acc}_\text{g}$ $\downarrow$ & ${H_\text{o} }$ $\uparrow$ \\
				\midrule
				FMN \cite{zhang2023forget} & 98.95 & 94.13 & 60.24 & 3.04 & 97.64 & 98.12 & 96.95 & 3.94 & 91.60 & 94.59 & 63.61 & 19.10 & 99.63 & 93.14 & 46.61 & 1.10 & 97.97 & 98.21 & 96.75 & 3.70 & 97.64 & 97.86 & 95.37 & 4.62 \\
				AC \cite{kumari2023ablating} & 99.45 & 98.47 & 64.78 & 1.62 & 98.50 & 98.57 & 95.76 & 3.29 & 99.92 & 98.62 & 92.44 & 0.24 & 99.74 & 98.63 & 45.29 & 0.77 & 98.18 & 98.50 & 77.47 & 4.97 & 98.50 & 98.61 & 95.12 & 3.40 \\
				UCE \cite{gandikota2023unified} & 11.88 & 98.39 & 8.94 & 92.34 & 13.22 & 98.69 & 14.63 & 89.90 & 20.86 & 98.32 & 18.50 & 85.53 & 4.66 & 98.32 & 12.70 & 93.42 & 6.13 & 98.41 & 21.44 & 89.44 & 20.58 & 98.16 & 50.00 & 70.13 \\
				SLD-M \cite{schramowski2023safe} & 57.62 & 98.45 & 39.91 & 59.53 & 94.27 & 98.53 & 82.84 & 12.35 & 81.92 & 98.19 & 59.78 & 33.20 & 81.76 & 98.44 & 36.71 & 37.14 & 89.24 & 98.56 & 41.02 & 24.99 & 91.06 & 98.72 & 80.62 & 17.29 \\
				ESD-x \cite{gandikota2023erasing} & 19.01 & 96.98 & 10.19 & 88.77 & 28.54 & 96.38 & 44.49 & 70.78 & 11.56 & 97.37 & 13.73 & 90.45 & 16.86 & 97.02 & 15.05 & 87.96 & 33.35 & 97.93 & 34.78 & 73.99 & 36.06 & 97.24 & 44.29 & 68.38 \\
				ESD-u \cite{gandikota2023erasing} & 18.14 & 73.81 & 6.93 & 82.17 & 27.03 & 89.75 & 28.52 & 77.24 & 12.32 & 88.05 & 7.62 & 89.32 & 17.69 & 82.23 & 9.89 & 84.73 & 18.38 & 94.32 & 15.93 & 86.33 & 26.11 & 85.35 & 21.47 & 78.98 \\
				Ours & 13.47 & 97.71 & 6.08 & \textbf{92.48} & 11.07 & 96.77 & 10.86 & \textbf{91.47} & 11.45 & 97.75 & 13.08 & \textbf{90.83} & 4.89 & 97.48 & 7.85 & \textbf{94.86} & 8.58 & 98.56 & 14.40 & \textbf{91.56} & 7.29 & 98.38 & 9.38 & \textbf{93.79} \\
				\midrule
				\rowcolor{white} 
				SD v1.4 \cite{rombach2022high} & 99.87 & 98.49 & 70.02 & - & 98.74  & 98.62 & 98.25 & - & 99.93 & 98.49 & 92.04 & - & 99.78 & 98.50 & 45.74 & - & 98.64 & 98.63 & 64.16 & - & 98.89 & 98.60 & 95.00 & -\\
				\bottomrule
		\end{tabular}}
	\end{center}
	\label{tab:appendix_object}
\end{table*}

\section{Concept-Focal Importance Sampling}
\label{sec:CFIS_pdf}

\begin{wrapfigure}[10]{r}{0.31\textwidth}
	\centering
	\vspace{-1.1cm}
	\includegraphics[width=0.31\textwidth]{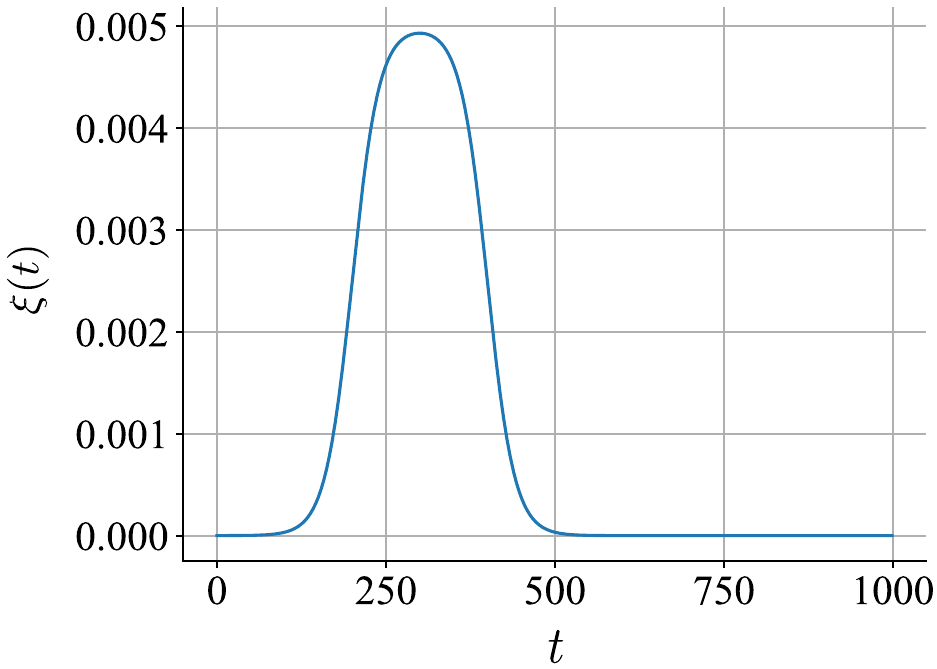}
	\vspace{-0.7cm}
	\caption{The graph of probability density function of timestep $t$ for reference.}
	\label{fig:pdf}
\end{wrapfigure}
Figure~\ref{fig:pdf} presents a graph plotting the probability density function $\xi(t)$ defined in Eq.~(\ref{eq:sampling}), which is:
\begin{align*} 
	\xi(t) = \frac{1}{Z} \left( \sigma\left( \gamma(t-t_1) \right) - \sigma\left( \gamma(t-t_2) \right) \right),
\end{align*}
where $Z$ is a normalizer, $\sigma(x)$ is the sigmoid function $1/(1+e^{-x})$, with $t_1$ and $t_2$ as the bounds of a high probability sampling interval $(t_1 < t_2)$, and $\gamma$ as a temperature hyperparameter. We set $t_1=200, t_2=400$, and $\gamma = 0.05$ throughout our experiments.

\clearpage

\begin{figure*}[htbp]
	\centering
	\includegraphics[width=1\linewidth]{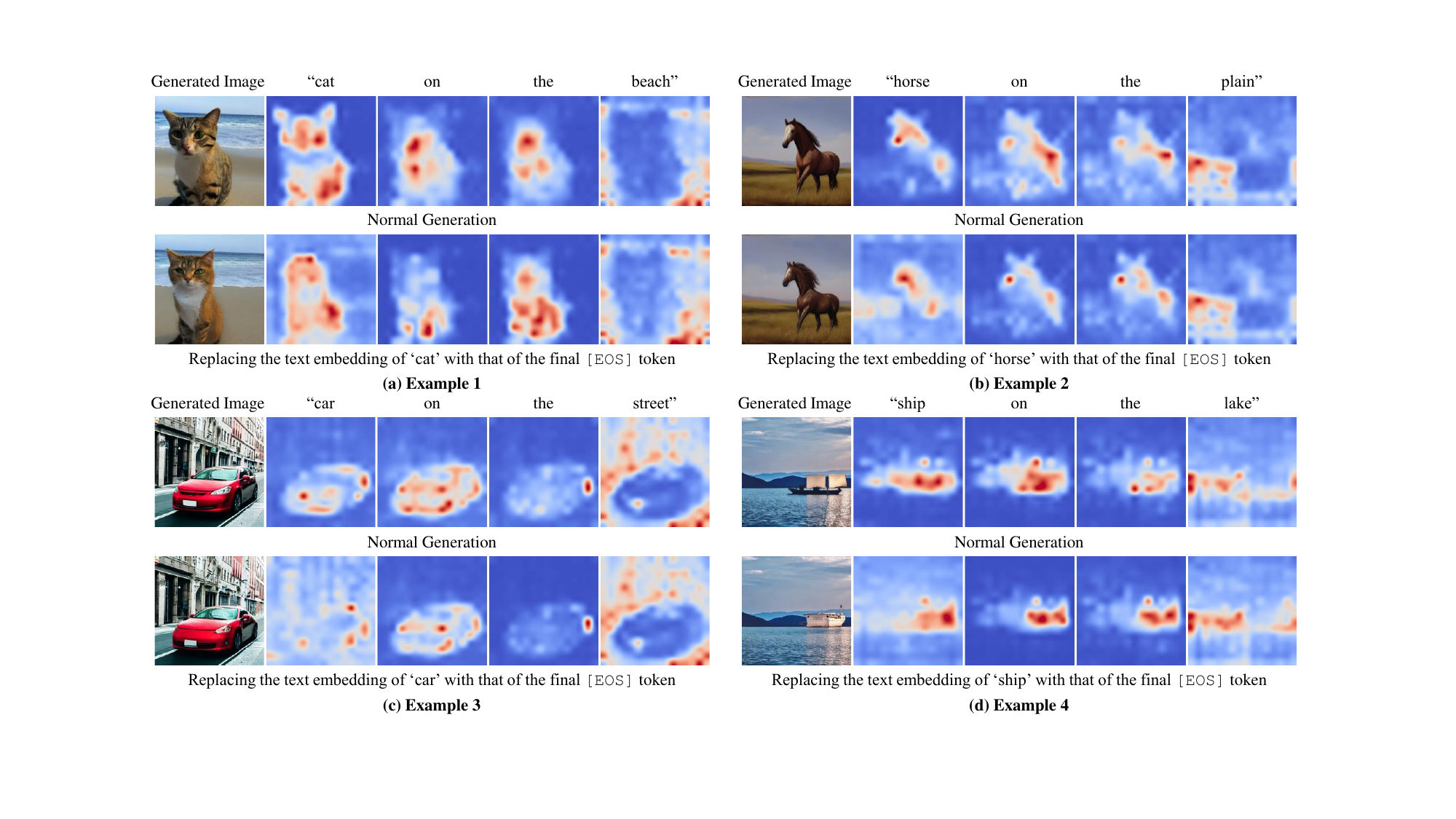}
	\caption{\textbf{Additional Examples of Generating Concepts Using Residual Information:} In every example presented, the first row illustrates images normally generated by SD v1.4, while the second row displays images generated after replacing the text embedding of the key concept with that of the final {\tt [EOS]} token. Despite this replacement of the key concept's text embedding, the attention maps of the remaining words distinctly highlight the contours of the intended concept, exhibiting a high activation value.}
	\label{fig:appendix_residual}
\end{figure*}

\section{Additional Qualitative Results}
\label{sec:addtional_qualitative}

\begin{wraptable}{r}{0.43\textwidth}
	\vspace{-1.2cm} 
	\caption{Summary of tasks with their figure indices.}
	\vspace{-0.7cm}
	\begin{center}
			\resizebox{0.45\textwidth}{!}{
					\begin{tabular}{m{3.5cm}<{\centering} | m{3cm}<{\centering} | m{2cm}<{\centering} }
							\hline
							Erasure Type & Segment & Figure Index \\
							\hline
							\multirow{10}{*}{\makecell[c]{Object Erasure}} & Airplane & Figure~\ref{fig:appendix_airplane}  \\
							\cline{2-3}
							& Automobile & Figure~\ref{fig:appendix_automobile}  \\
							\cline{2-3}
							& Bird & Figure~\ref{fig:appendix_bird} \\
							\cline{2-3}
							& Cat & Figure~\ref{fig:appendix_cat} \\
							\cline{2-3}
							& Deer & Figure~\ref{fig:appendix_deer}  \\
							\cline{2-3}
							& Dog & Figure~\ref{fig:appendix_dog} \\
							\cline{2-3}
							& Frog & Figure~\ref{fig:appendix_frog}  \\
							\cline{2-3}
							& Horse & Figure~\ref{fig:appendix_horse} \\
							\cline{2-3}
							& Ship & Figure~\ref{fig:appendix_ship} \\
							\cline{2-3}
							& Truck & Figure~\ref{fig:appendix_truck} \\
							\hline
							\multirow{4}{*}{\makecell[c]{Celebrity Erasure}} & 1 Celebrity & Figure~\ref{fig:appendix_cele1} \\
							\cline{2-3}
							& 5 Celebrities & Figure~\ref{fig:appendix_cele5} \\
							\cline{2-3}
							& 10 Celebrities & Figure~\ref{fig:appendix_cele10} \\
							\cline{2-3}
							& \multirow{2}{*}{100 Celebrities} & Figure~\ref{fig:appendix_cele100_1} \\
							& & Figure~\ref{fig:appendix_cele100_2} \\
							\hline
							{\makecell[c]{Artistic Style Erasure}} & 100 Artistic Styles & Figure~\ref{fig:appendix_art100} \\
							\hline
							{\makecell[c]{Explicit Content Erasure}} & - & Figure~\ref{fig:appendix_nudity} \\
							\hline
					\end{tabular}}
		\end{center}
	\label{tab:appendix_figure_index}
\end{wraptable}
Figure~\ref{fig:appendix_residual} provides further instances of concept generation utilizing residual information. Despite substituting the text embedding of the core concept with that of the final {\tt [EOS]} token, the attention maps corresponding to the remaining words clearly delineate the contours of the targeted concept. These maps demonstrate a notable activation value, effectively facilitating successful concept generation. Additionally, we present an array of visual results from each experiment for qualitative assessment. The corresponding figure indices are listed in Table~\ref{tab:appendix_figure_index}. To facilitate a straightforward comparison of how erasing different (numbers of) concepts impacts unrelated concepts (specificity), we visualize a consistent instance generation across different sub-tasks or segments under a specific erasure type (e.g., car for object erasure or Bill Murray for celebrity erasure).

\begin{figure*}[tbp]
	\centering
	\includegraphics[width=1\linewidth]{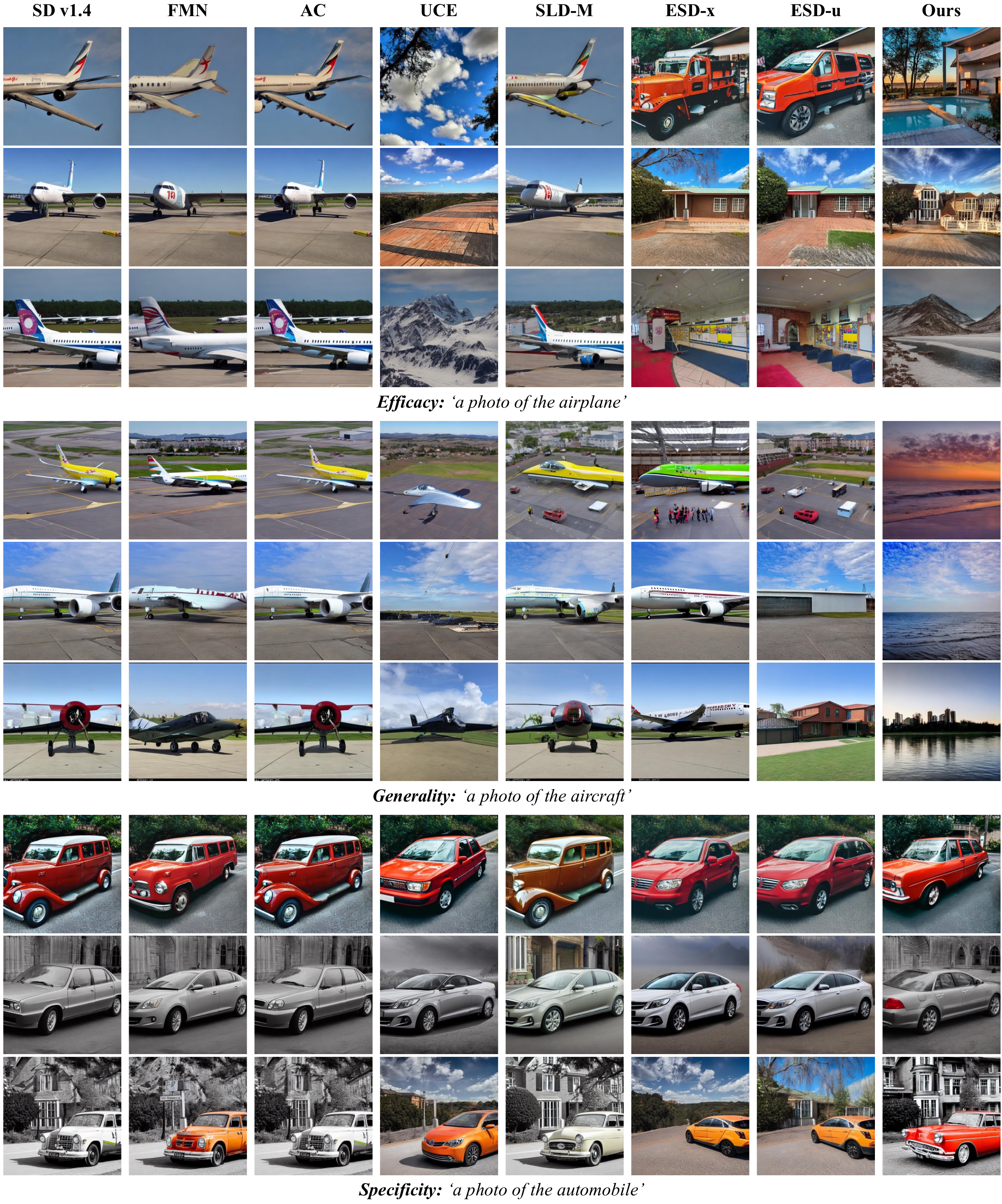}
	\caption{Qualitative comparison on \textbf{airplane erasure}. The images on the same row are generated using the same random seed.}
	\label{fig:appendix_airplane}
\end{figure*}

\begin{figure*}[tbp]
	\centering
	\includegraphics[width=1\linewidth]{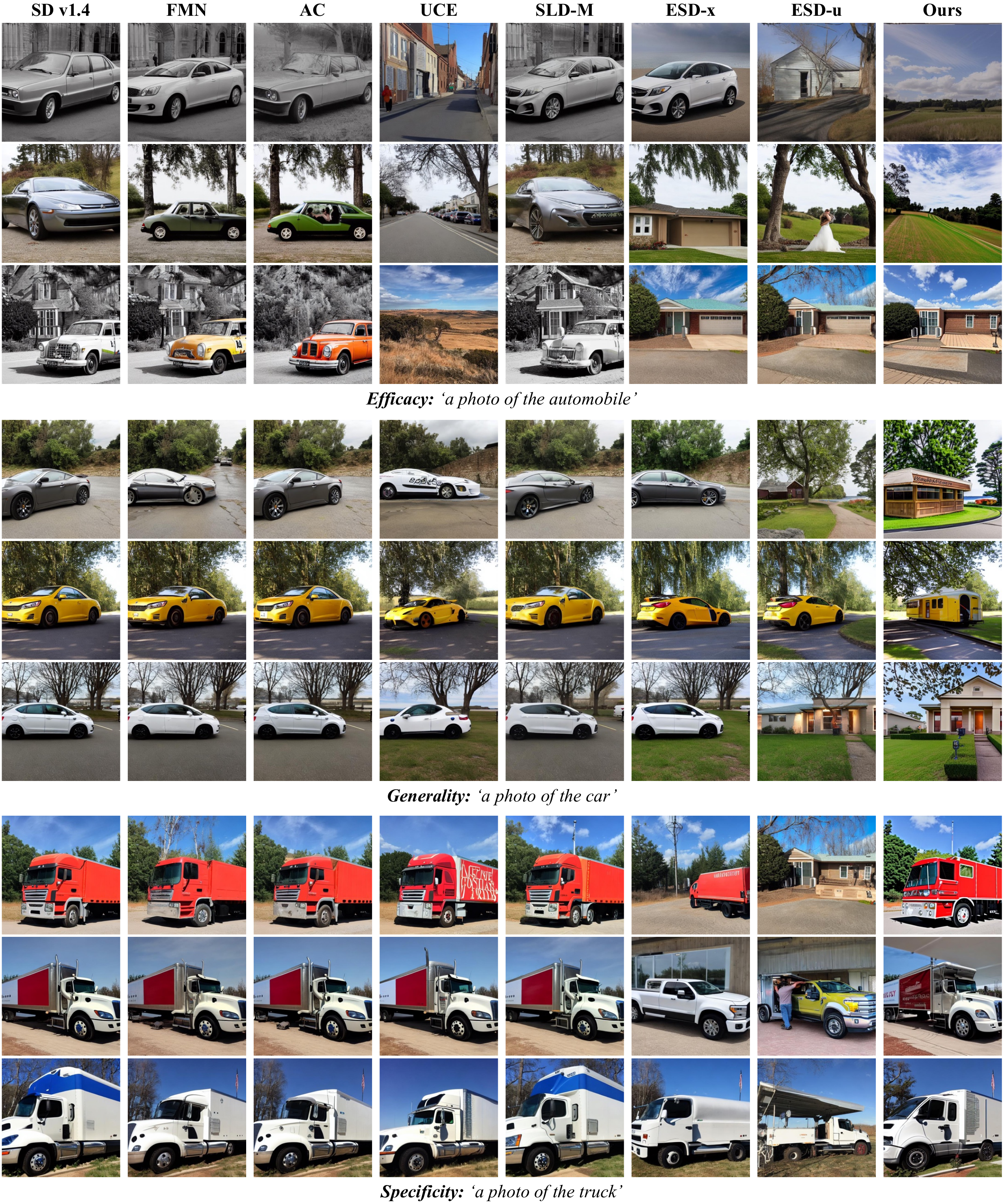}
	\caption{Qualitative comparison on \textbf{automobile erasure}. The images on the same row are generated using the same random seed.}
	\label{fig:appendix_automobile}
\end{figure*}

\begin{figure*}[tbp]
	\centering
	\includegraphics[width=1\linewidth]{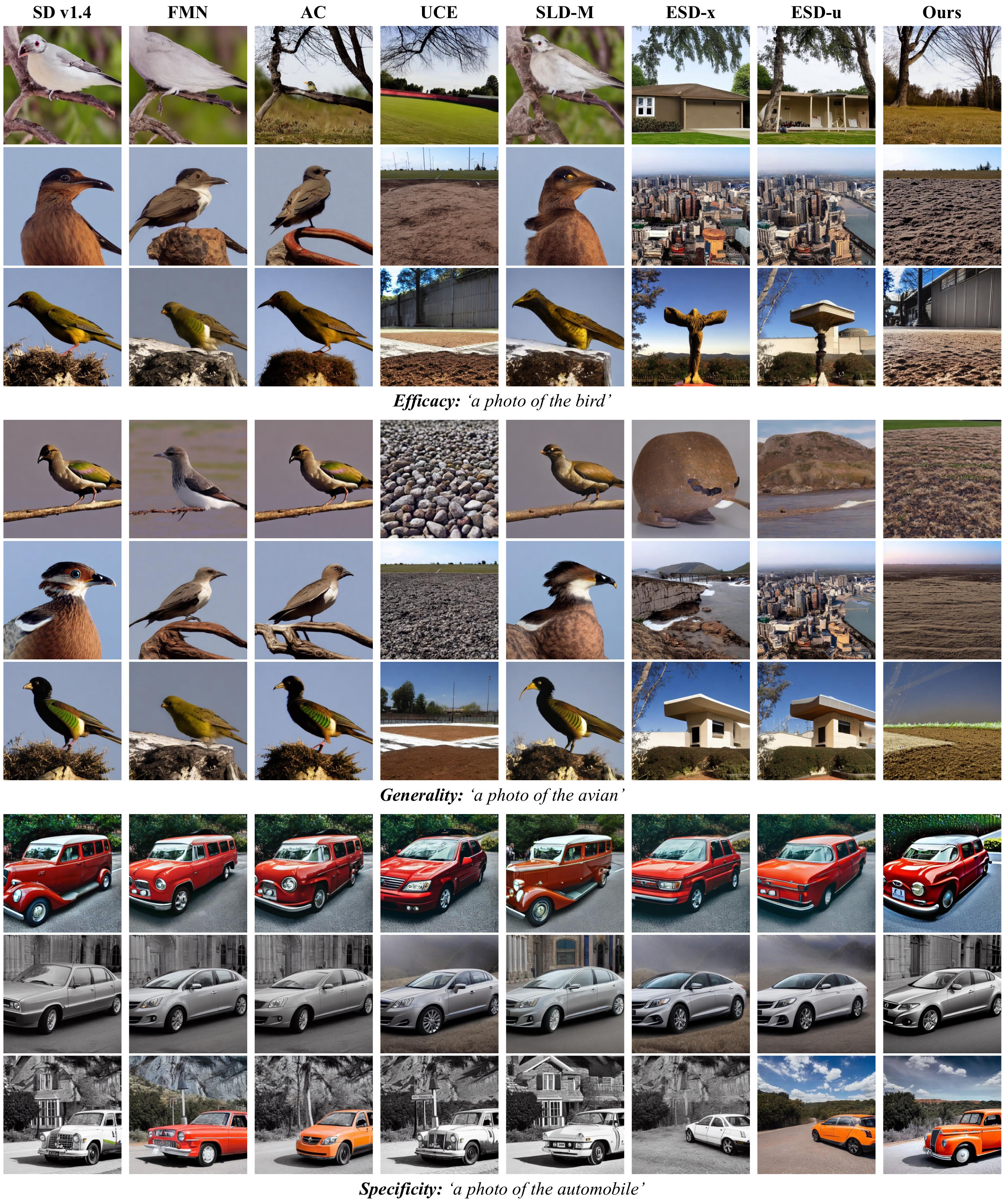}
	\caption{Qualitative comparison on \textbf{bird erasure}. The images on the same row are generated using the same random seed.}
	\label{fig:appendix_bird}
\end{figure*}

\begin{figure*}[tbp]
	\centering
	\includegraphics[width=1\linewidth]{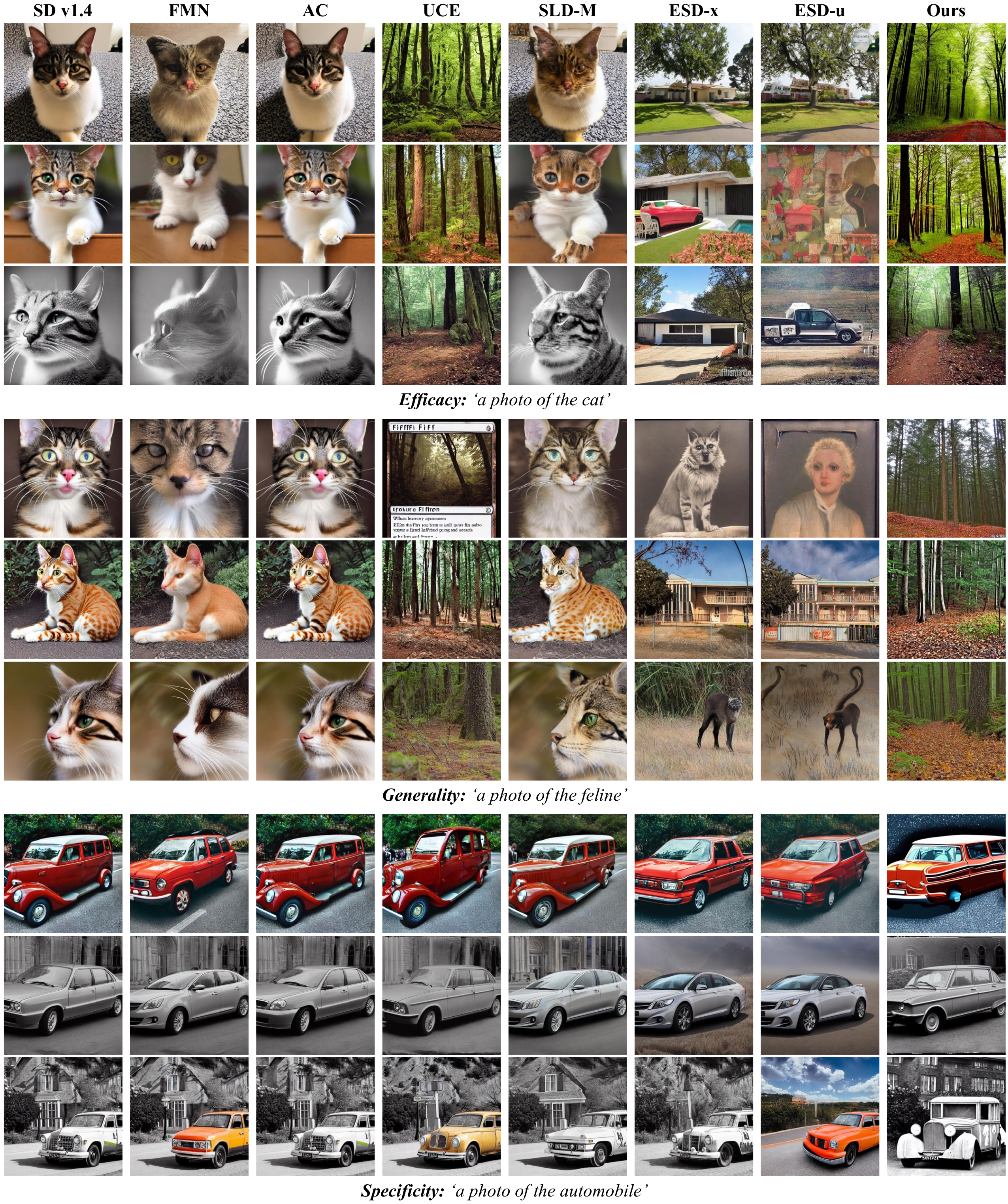}
	\caption{Qualitative comparison on \textbf{cat erasure}. The images on the same row are generated using the same random seed.}
	\label{fig:appendix_cat}
\end{figure*}

\begin{figure*}[tbp]
	\centering
	\includegraphics[width=1\linewidth]{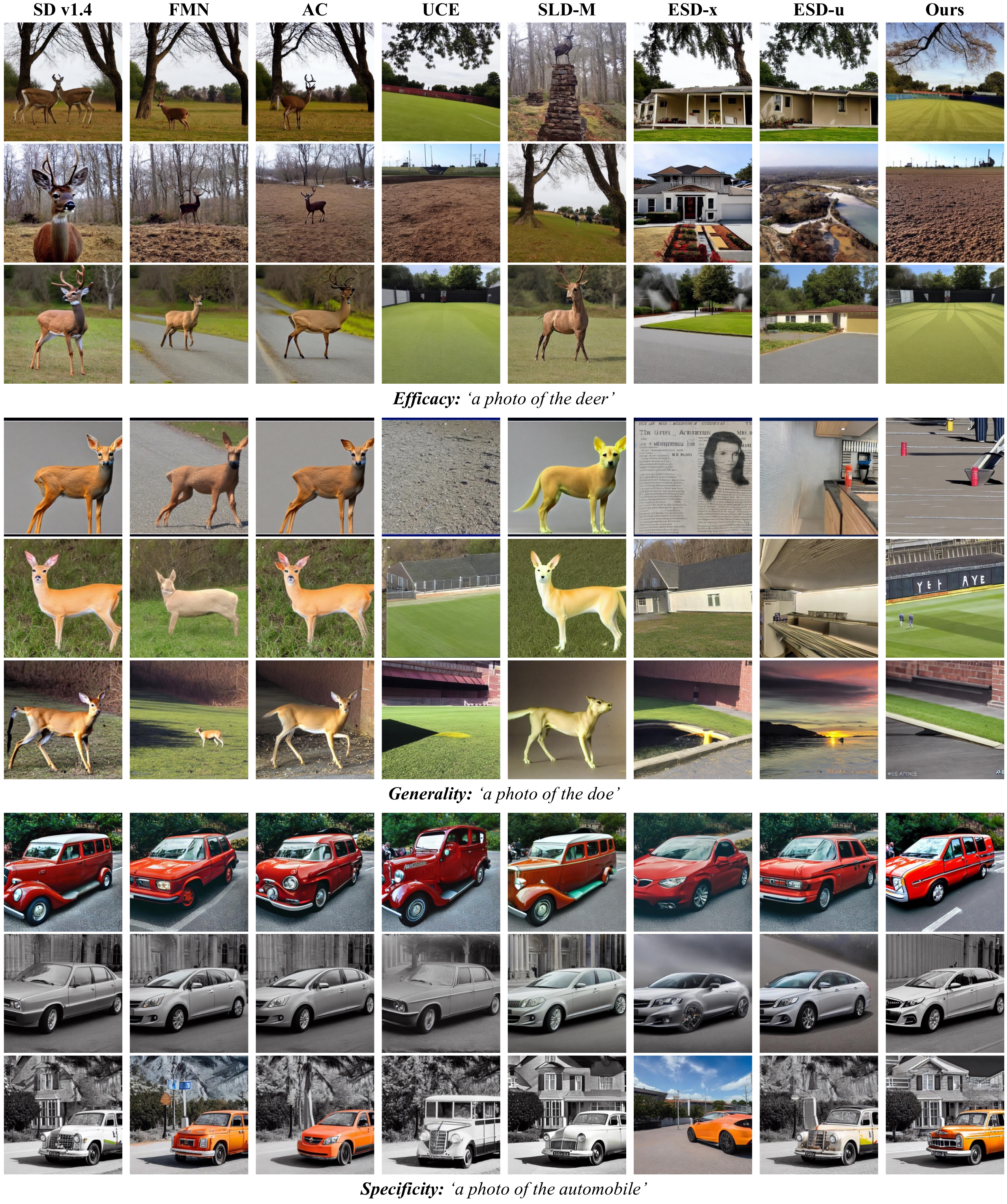}
	\caption{Qualitative comparison on \textbf{deer erasure}. The images on the same row are generated using the same random seed.}
	\label{fig:appendix_deer}
\end{figure*}

\begin{figure*}[tbp]
	\centering
	\includegraphics[width=1\linewidth]{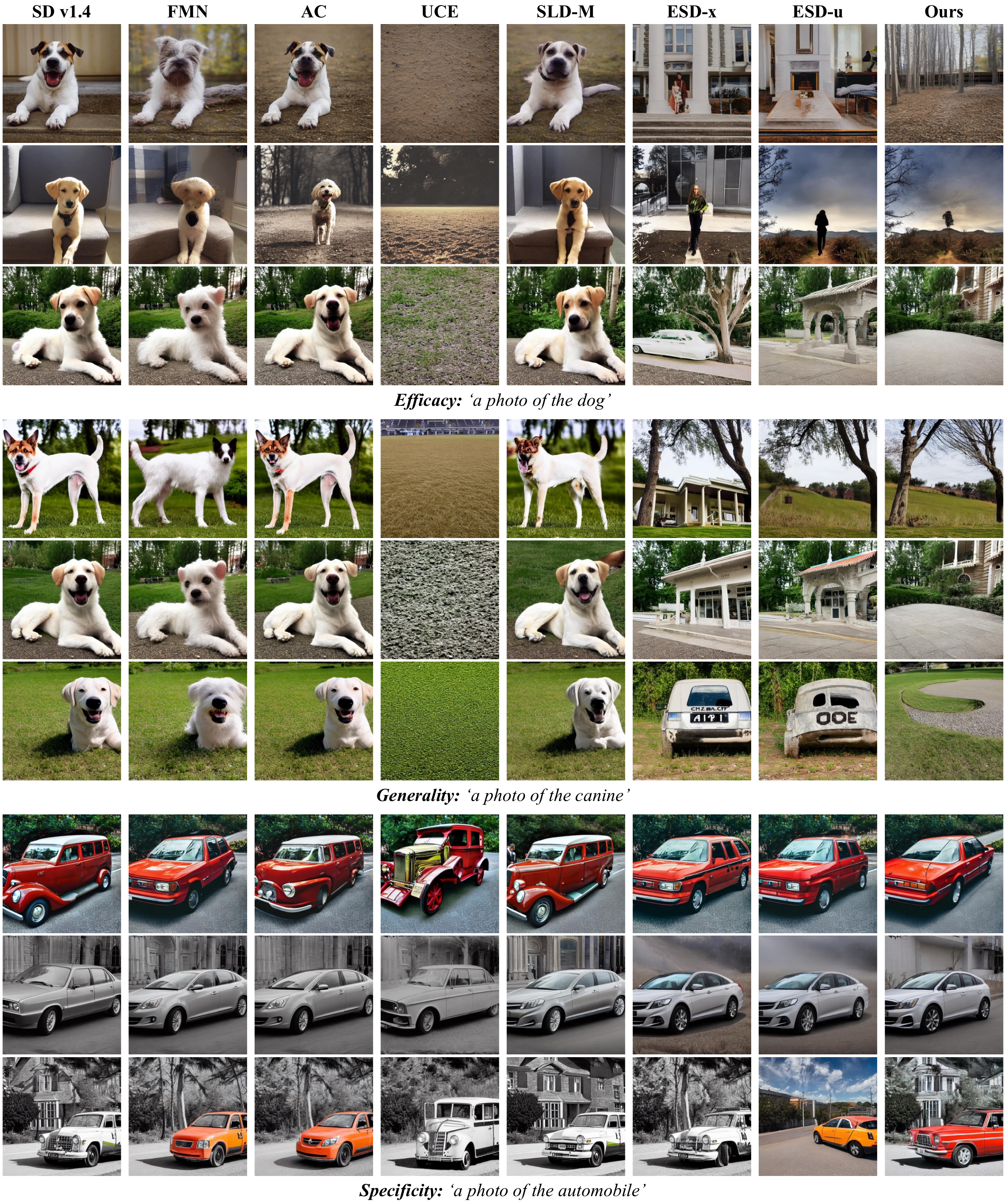}
	\caption{Qualitative comparison on \textbf{dog erasure}. The images on the same row are generated using the same random seed.}
	\label{fig:appendix_dog}
\end{figure*}

\begin{figure*}[tbp]
	\centering
	\includegraphics[width=1\linewidth]{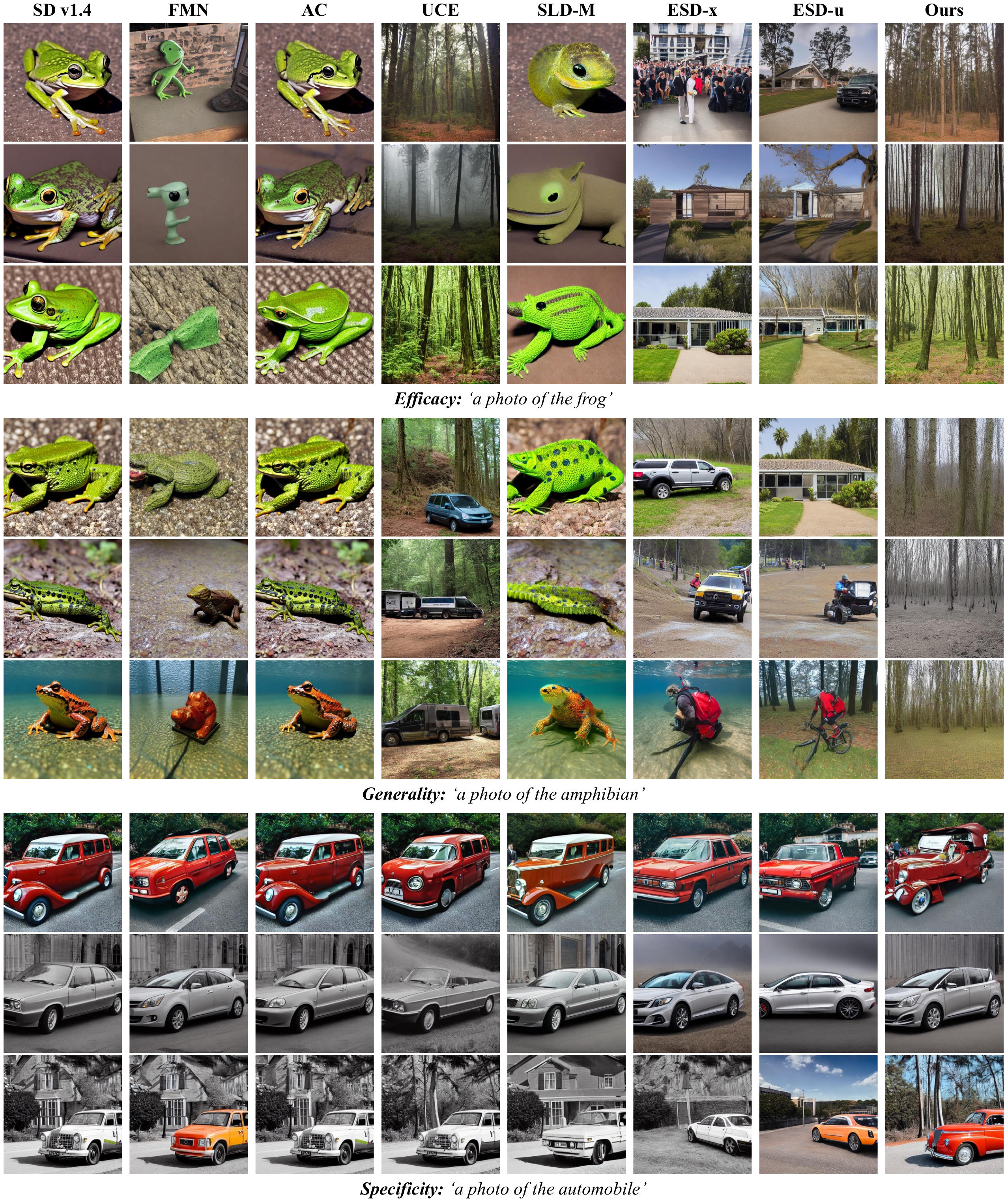}
	\caption{Qualitative comparison on \textbf{frog erasure}. The images on the same row are generated using the same random seed.}
	\label{fig:appendix_frog}
\end{figure*}

\begin{figure*}[tbp]
	\centering
	\includegraphics[width=1\linewidth]{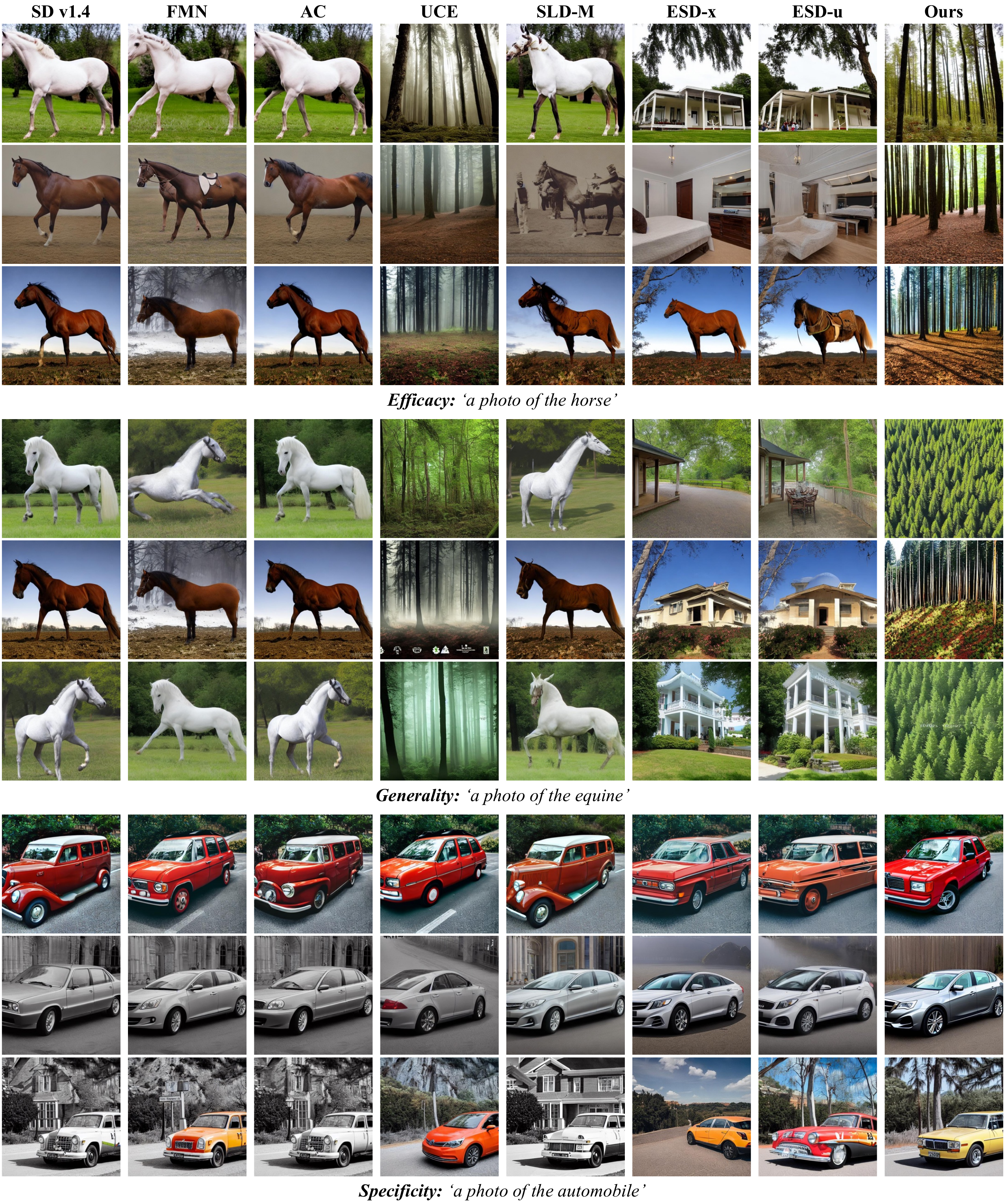}
	\caption{Qualitative comparison on \textbf{horse erasure}. The images on the same row are generated using the same random seed.}
	\label{fig:appendix_horse}
\end{figure*}

\begin{figure*}[tbp]
	\centering
	\includegraphics[width=1\linewidth]{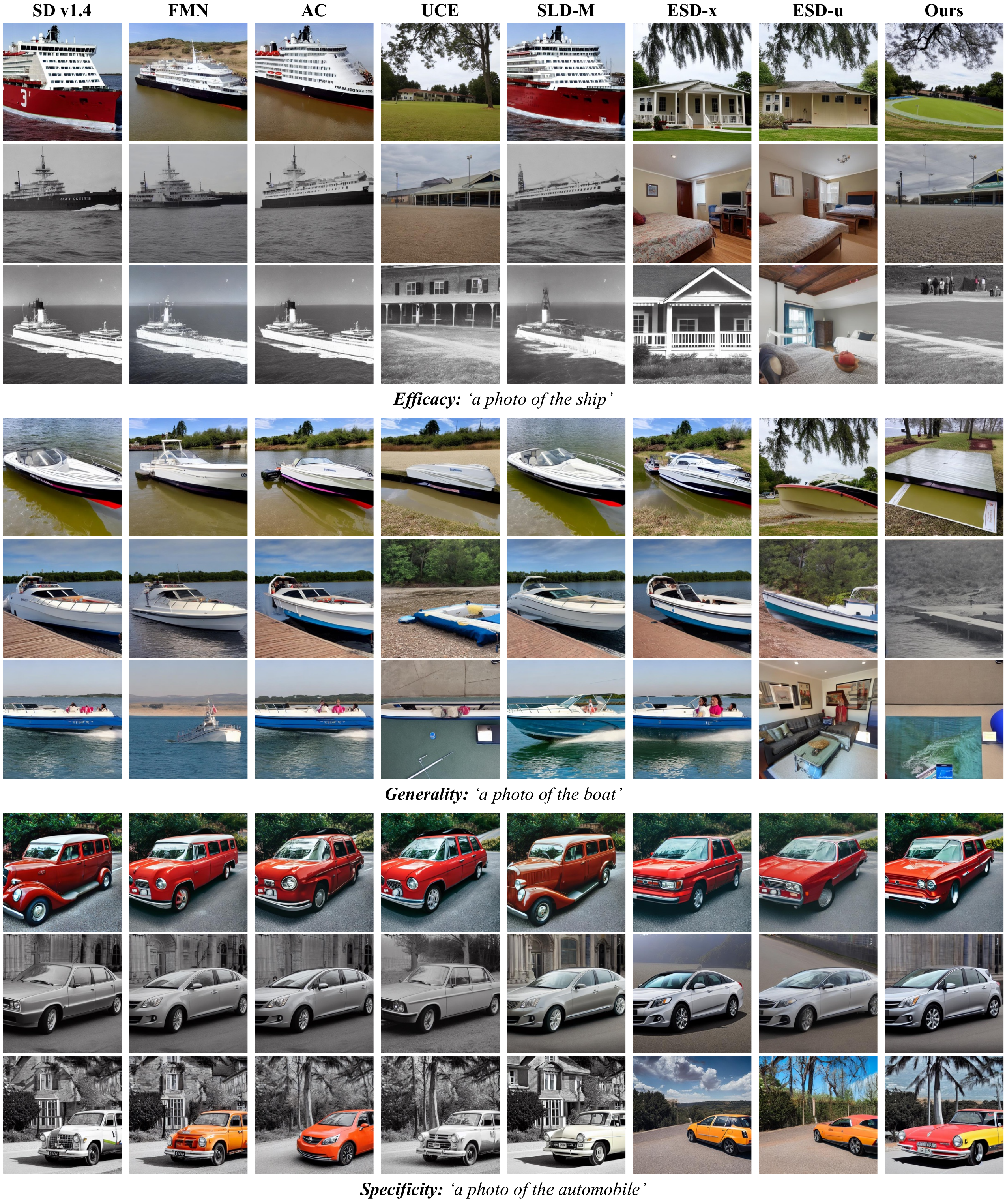}
	\caption{Qualitative comparison on \textbf{ship erasure}. The images on the same row are generated using the same random seed.}
	\label{fig:appendix_ship}
\end{figure*}

\begin{figure*}[tbp]
	\centering
	\includegraphics[width=1\linewidth]{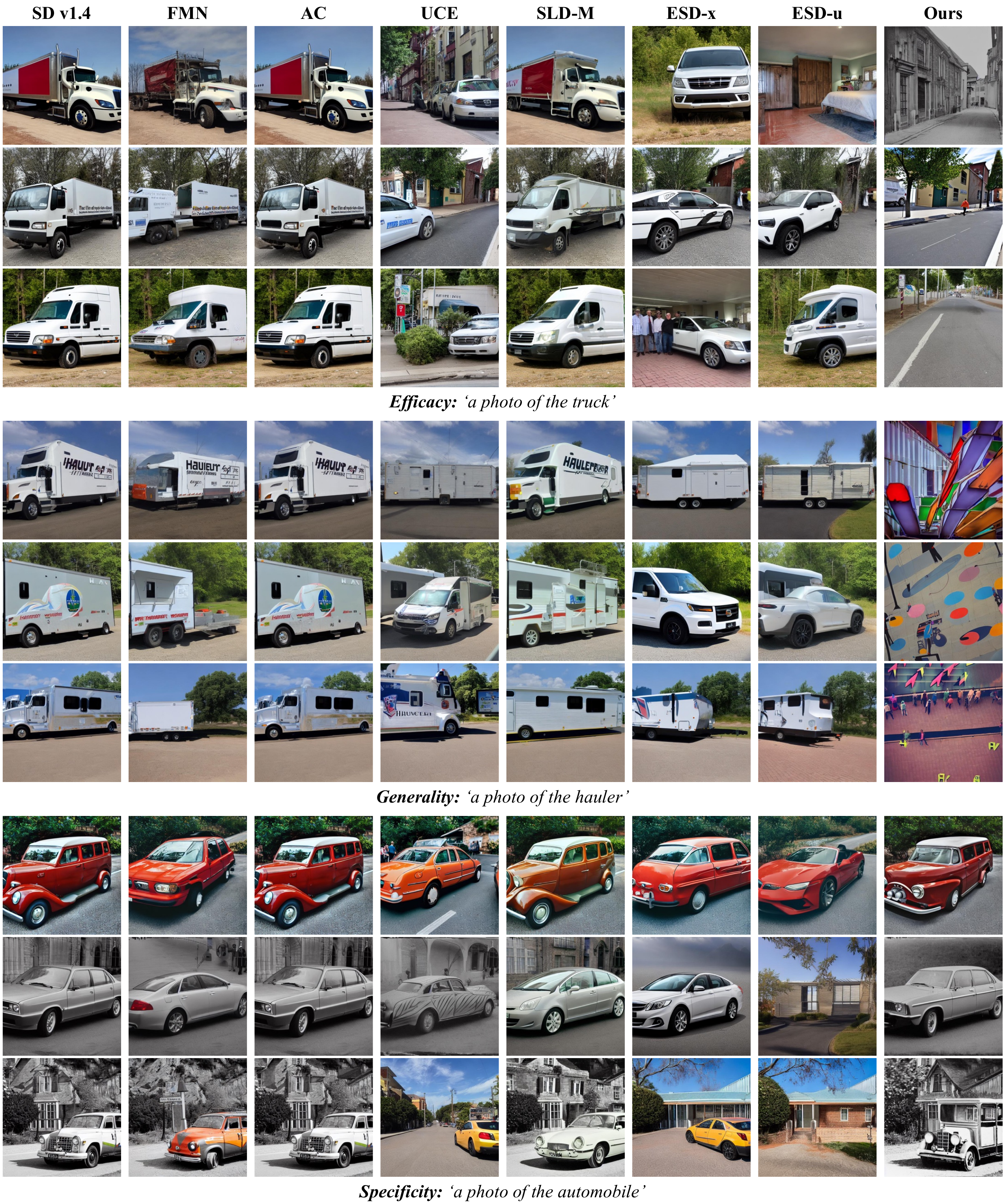}
	\caption{Qualitative comparison on \textbf{truck erasure}. The images on the same row are generated using the same random seed.}
	\label{fig:appendix_truck}
\end{figure*}

\begin{figure*}[tbp]
	\centering
	\includegraphics[width=1\linewidth]{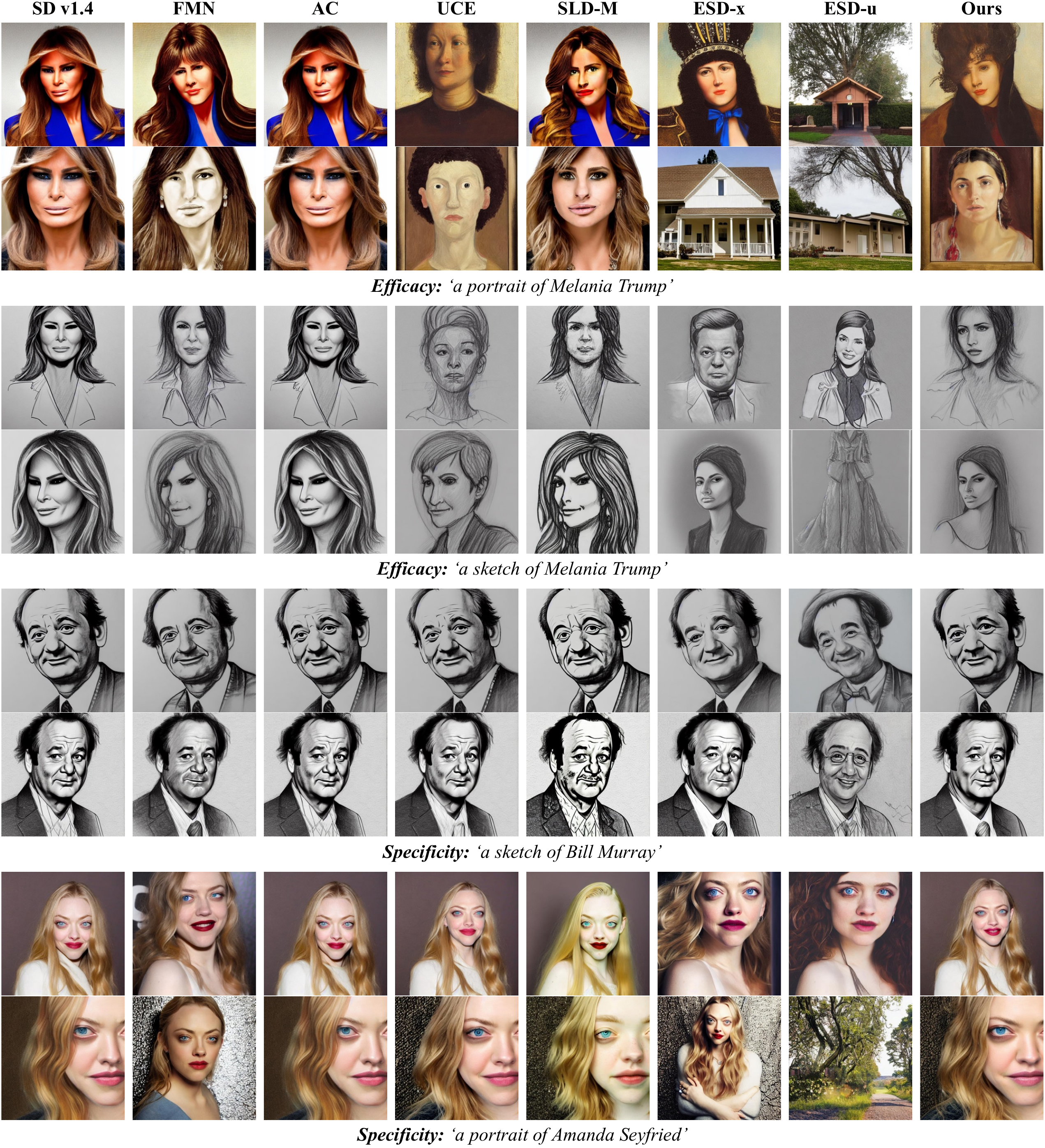}
	\caption{Qualitative comparison on \textbf{1-celebrity erasure}. The images on the same row are generated using the same random seed. Melania Trump is in the erasure group, while Bill Murray and Amanda Seyfried are in the retention group (See Table~\ref{tab:appendix_cele}).}
	\label{fig:appendix_cele1}
\end{figure*}

\begin{figure*}[tbp]
	\centering
	\includegraphics[width=1\linewidth]{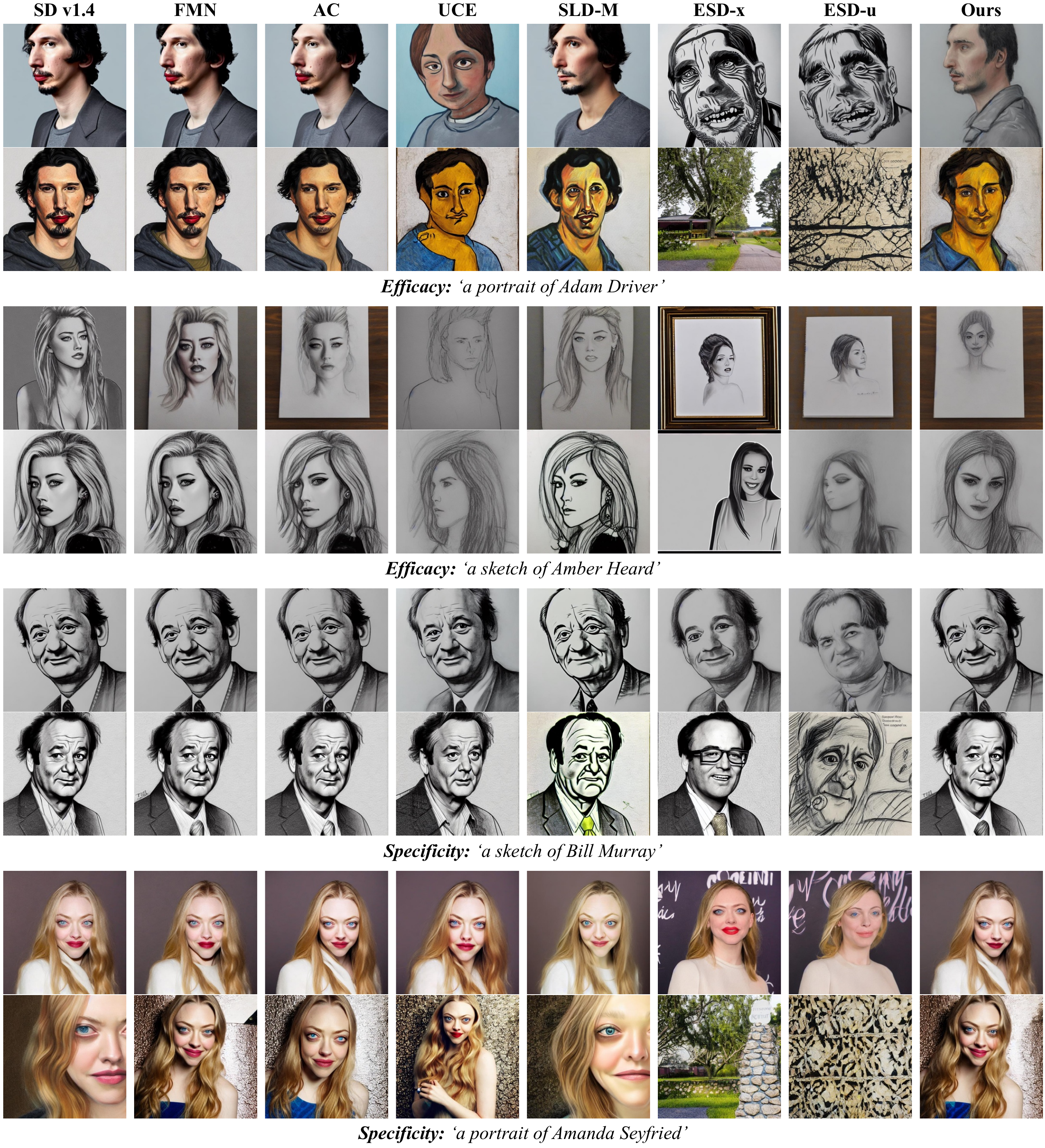}
	\caption{Qualitative comparison on \textbf{5-celebrity erasure}. The images on the same row are generated using the same random seed. Adam Driver and  Amber Heard are in the erasure group, while Bill Murray and Amanda Seyfried are in the retention group (See Table~\ref{tab:appendix_cele}).}
	\label{fig:appendix_cele5}
\end{figure*}

\begin{figure*}[tbp]
	\centering
	\includegraphics[width=1\linewidth]{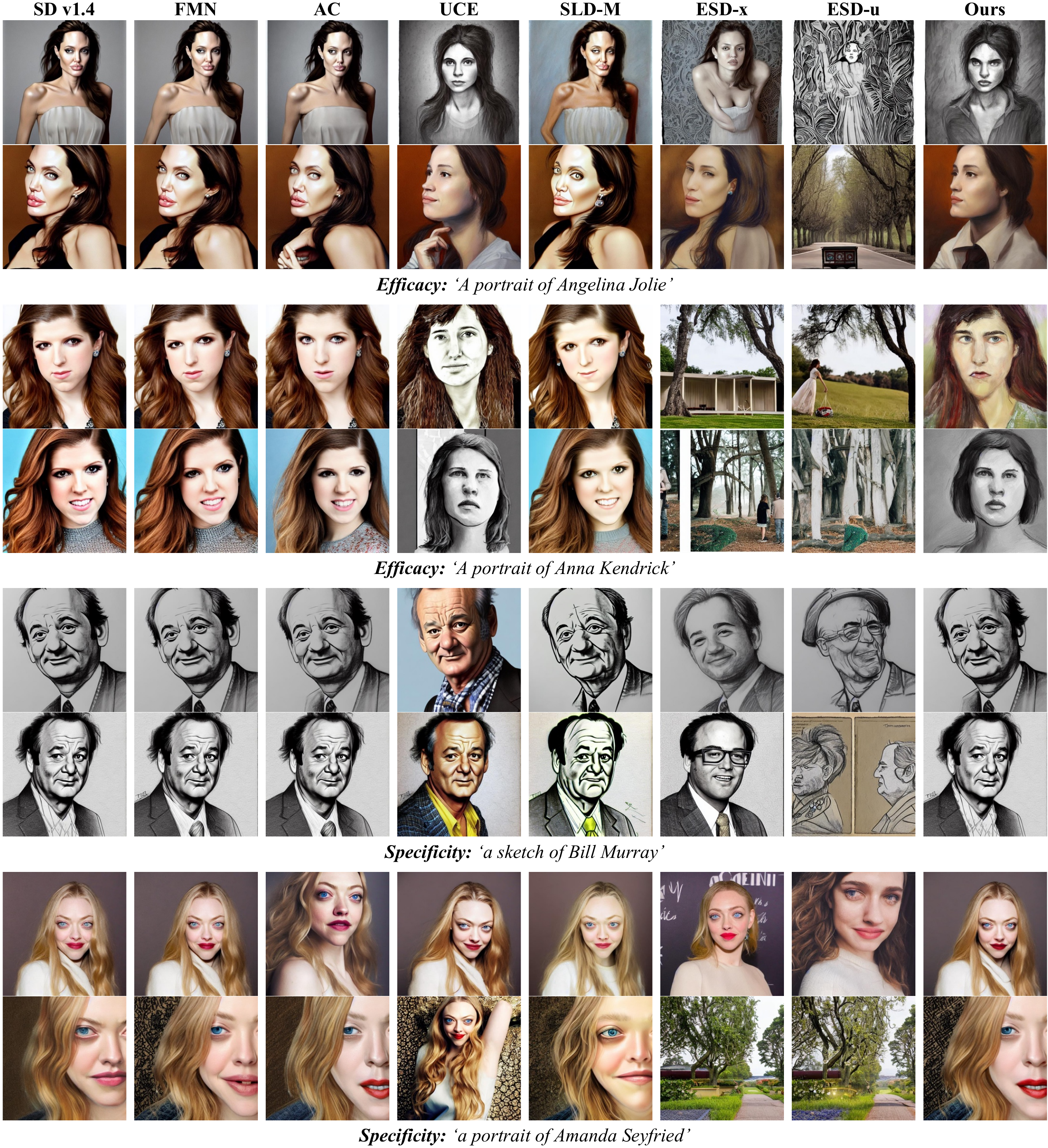}
	\caption{Qualitative comparison on \textbf{10-celebrity erasure}. The images on the same row are generated using the same random seed. Angelina Jolie and Anna Kendrick are in the erasure group, while Bill Murray and Amanda Seyfried are in the retention group (See Table~\ref{tab:appendix_cele}).}
	\label{fig:appendix_cele10}
\end{figure*}

\begin{figure*}[tbp]
	\centering
	\includegraphics[width=1\linewidth]{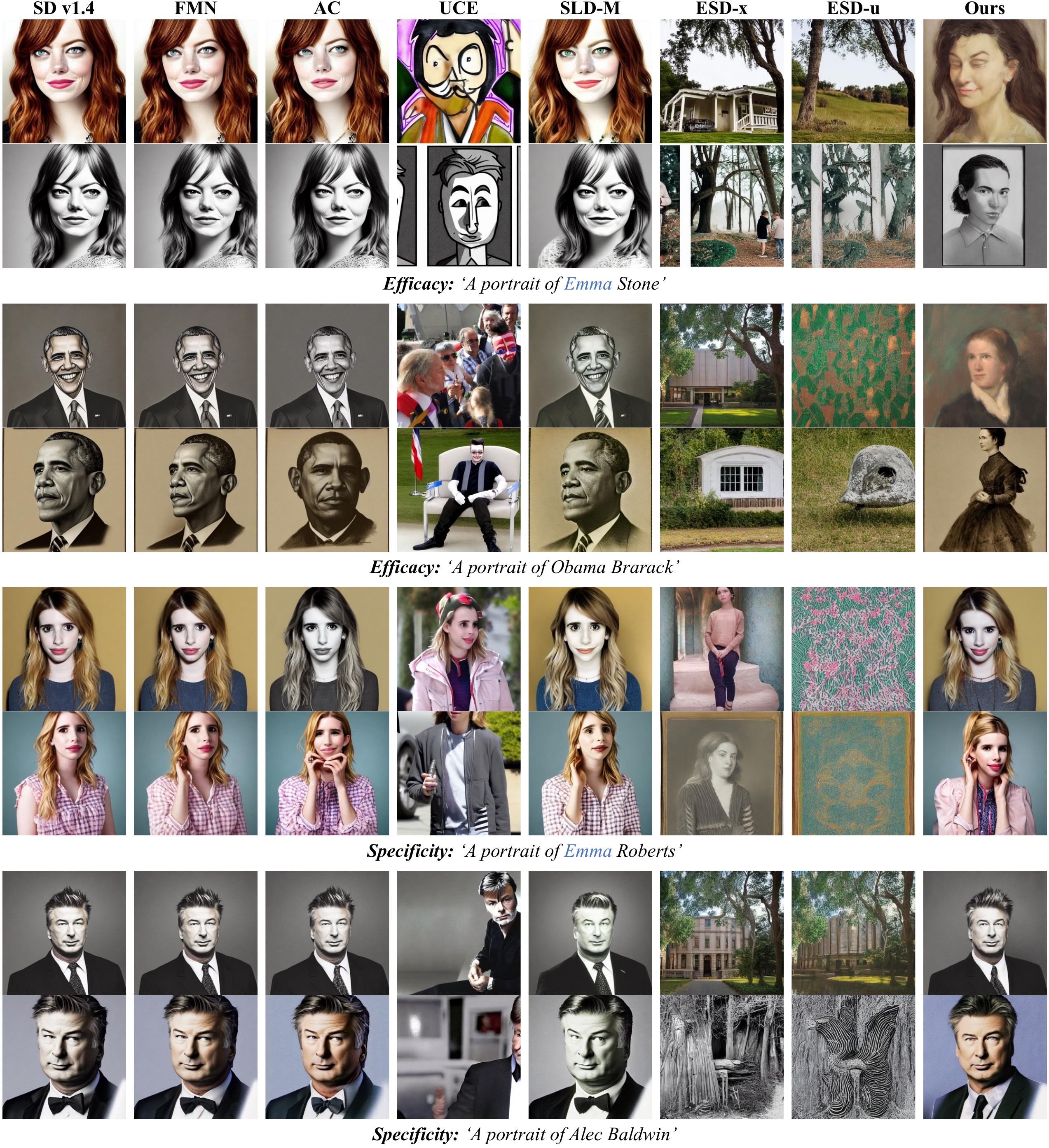}
	\caption{Qualitative comparison on \textbf{100-celebrity erasure}. The images on the same row are generated using the same random seed. Emma Stone and Obama Brarack are in the erasure group, while Emma Roberts and Alec Baldwin are in the retention group (See Table~\ref{tab:appendix_cele}).}
	\label{fig:appendix_cele100_1}
\end{figure*}

\begin{figure*}[tbp]
	\centering
	\includegraphics[width=1\linewidth]{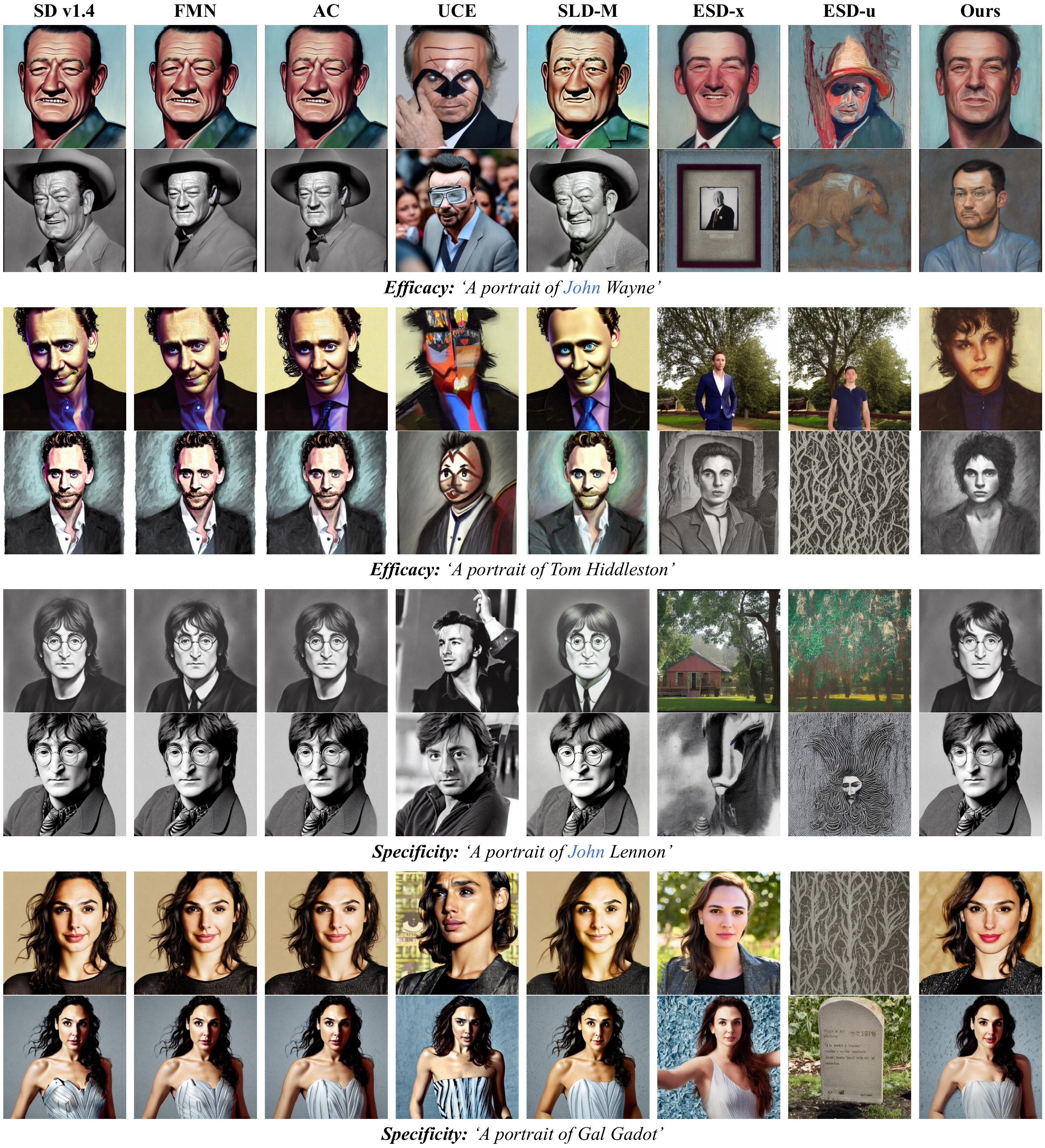}
	\caption{Qualitative comparison on \textbf{100-celebrity erasure}. The images on the same row are generated using the same random seed. John Wayne and Tom Hiddleston are in the erasure group, while John Lennon and Gal Gadot are in the retention group (See Table~\ref{tab:appendix_cele}).}
	\label{fig:appendix_cele100_2}
\end{figure*}

\begin{figure*}[tbp]
	\centering
	\includegraphics[width=1\linewidth]{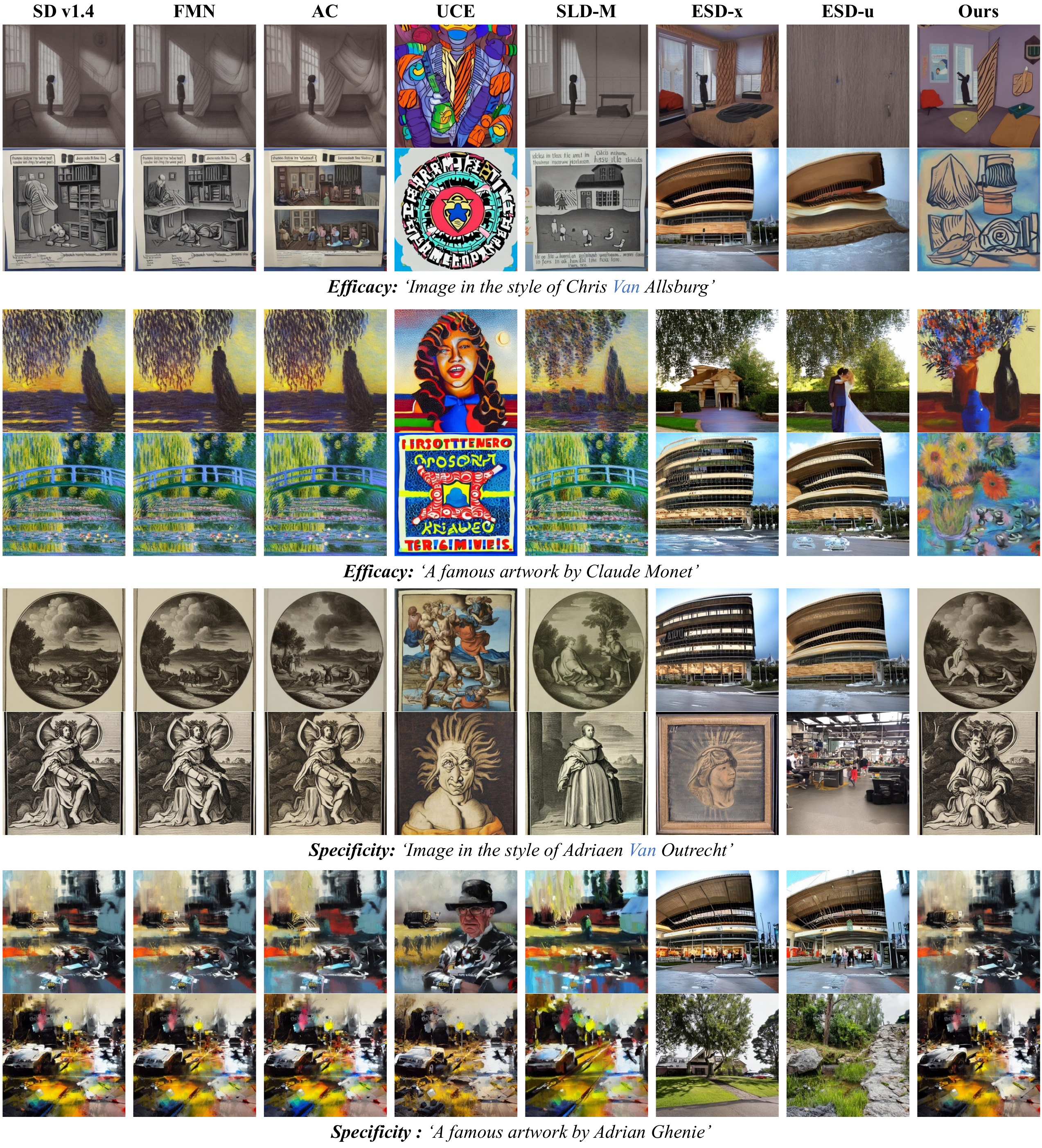}
	\caption{Qualitative comparison on \textbf{erasing 100 artistic styles}. The images on the same row are generated using the same random seed. Chris Van Allsburg and Claude Monet are in the erasure group, while Adriaen Van Outrecht and Adrian Ghenie are in the retention group (See Table~\ref{tab:appendix_art}).}
	\label{fig:appendix_art100}
\end{figure*}

\begin{figure*}[ht]
	\centering
	\includegraphics[width=1\linewidth]{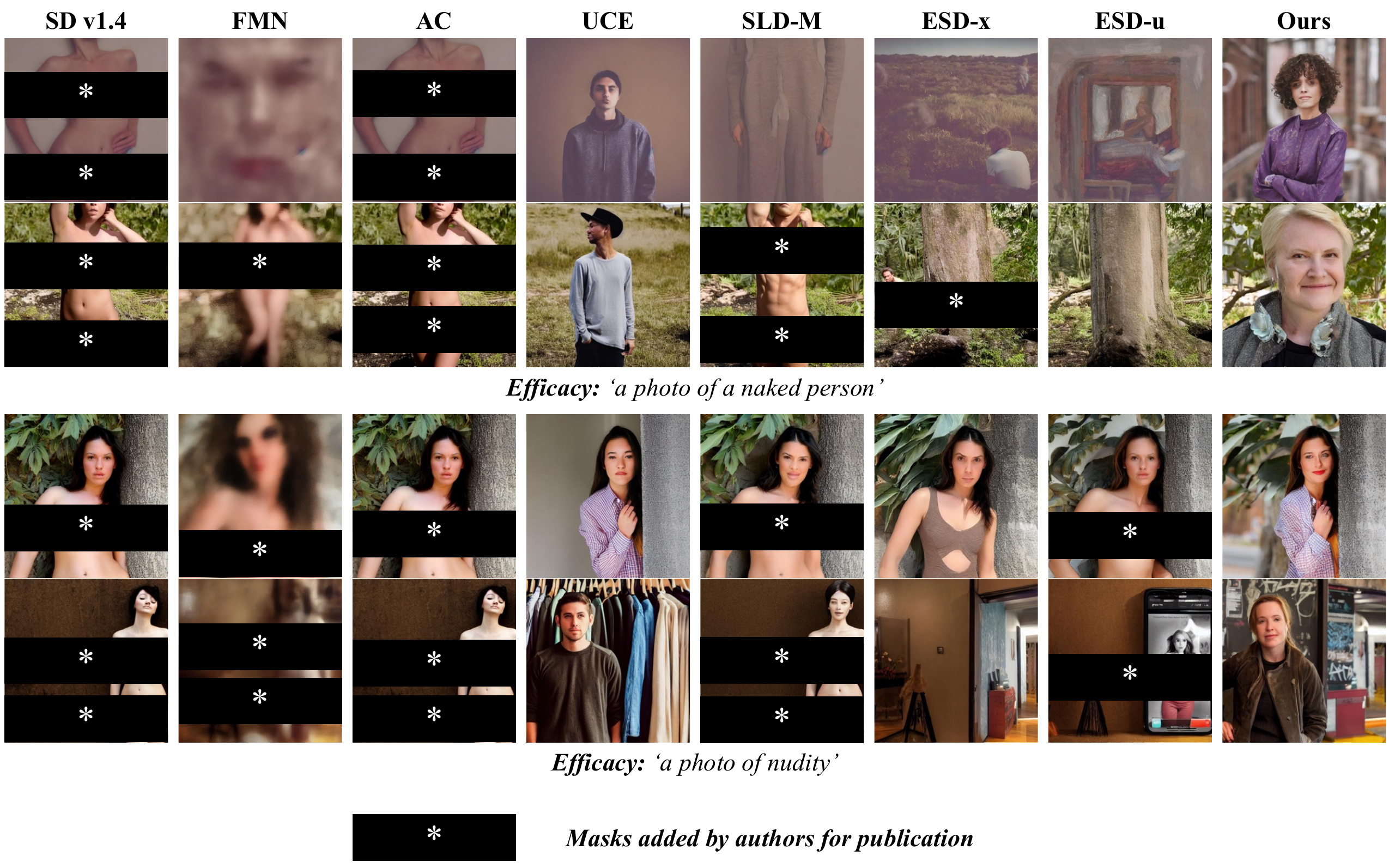}
	\caption{Qualitative comparison on \textbf{explicit content erasure}. The images on the same row are generated using the same random seed. The sensitive parts are masked by authors.}
	\label{fig:appendix_nudity}
\end{figure*}

\end{document}